# Approximate Model-Based Diagnosis
# Using Greedy Stochastic Search

**Alexander Feldman**                                        A.B.FELDMAN@TUDELFT.NL
*Delft University of Technology*
*Mekelweg 4, 2628 CD, Delft, The Netherlands*

**Gregory Provan**                                          G.PROVAN@CS.UCC.IE
*University College Cork*
*College Road, Cork, Ireland*

**Arjan van Gemund**                                        A.J.C.VANGEMUND@TUDELFT.NL
*Delft University of Technology*
*Mekelweg 4, 2628 CD, Delft, The Netherlands*

## Abstract

We propose a StochAstic Fault diagnosis AlgoRIthm, called SAFARI, which trades off guarantees of computing minimal diagnoses for computational efficiency. We empirically demonstrate, using the 74XXX and ISCAS85 suites of benchmark combinatorial circuits, that SAFARI achieves several orders-of-magnitude speedup over two well-known deterministic algorithms, CDA* and HA*, for multiple-fault diagnoses; further, SAFARI can compute a range of multiple-fault diagnoses that CDA* and HA* cannot. We also prove that SAFARI is optimal for a range of propositional fault models, such as the widely-used weak-fault models (models with ignorance of abnormal behavior). We discuss the optimality of SAFARI in a class of strong-fault circuit models with stuck-at failure modes. By modeling the algorithm itself as a Markov chain, we provide exact bounds on the minimality of the diagnosis computed. SAFARI also displays strong anytime behavior, and will return a diagnosis after any non-trivial inference time.

## 1. Introduction

Model-Based Diagnosis (MBD) is an area of artificial intelligence that uses a system model, together with observations about system behavior, to isolate sets of faulty components (diagnoses) that explain the observed behavior according to some minimality criterion. The standard MBD formalization (Reiter, 1987) frames a diagnostic problem in terms of a set of logical clauses that include mode-variables describing the nominal and fault status of system components; from this the diagnostic status of the system can be computed given an observation of the system's sensors. MBD provides a sound and complete approach to enumerating multiple-fault diagnoses, and exact algorithms can guarantee finding a diagnosis optimal with respect to the number of faulty components, probabilistic likelihood, etc.

The biggest challenge (and impediment to industrial deployment) is the computational complexity of the MBD problem. The MBD problem of determining if there exists a diagnosis with at most $k$ faults is NP-hard for the arbitrary propositional fault models we consider in this article (Bylander, Allemang, Tanner, & Josephson, 1991; Friedrich, Gottlob, & Nejdl, 1990). Computing the set of all diagnoses is harder still, since there are possibly exponen-tially many such diagnoses. Since almost all proposed MBD algorithms have been complete and exact, with some authors proposing possible trade-offs between completeness and faster consistency checking by employing methods such as BCP (Williams & Ragno, 2007), the complexity problem still remains a major challenge to MBD.

To overcome this complexity problem, we propose a novel *approximation approach* for multiple-fault diagnosis, based on a stochastic algorithm. Safari (StochAstic Fault diagnosis AlgoRIthm) sacrifices guarantees of optimality, but for diagnostic systems in which faults are described in terms of an arbitrary deviation from nominal behavior, Safari can compute diagnoses several orders of magnitude faster than competing algorithms.

Our contributions are as follows. (1) This paper introduces an approximation algorithm for computing diagnoses within an MBD framework, based on a greedy stochastic algorithm. (2) We show that we can compute minimal-cardinality diagnoses for weak fault models in polynomial time (calling an incomplete SAT-solver that implements Boolean Constraint Propagation[1] (BCP) only), and that more general frameworks (such as a sub-class of strong fault models) are also amenable to this class of algorithm. (3) We model Safari search as a Markov chain to show the performance and optimality trade-offs that the algorithm makes. (4) We apply this algorithm to a suite of benchmark combinatorial circuits, demonstrating order-of-magnitude speedup over two state-of-the-art deterministic algorithms, CDA* and HA*, for multiple-fault diagnoses. (5) We compare the performance of Safari against a range of Max-SAT algorithms for our benchmark problems. Our results indicate that, whereas the search complexity for the deterministic algorithms tested increases exponentially with fault cardinality, the search complexity for this stochastic algorithm appears to be independent of fault cardinality. Safari is of great practical significance, as it can compute a large fraction of minimal-cardinality diagnoses for discrete systems too large or complex to be diagnosed by existing deterministic algorithms.

## 2. Technical Background

Our discussion continues by formalizing some MBD notions. This paper uses the traditional diagnostic definitions (de Kleer & Williams, 1987), except that we use propositional logic terms (conjunctions of literals) instead of sets of failing components.

Central to MBD, a *model* of an artifact is represented as a propositional formula over some set of variables. Discerning two subsets of these variables as *assumable* and *observable*[2] variables gives us a diagnostic system.

**Definition 1** (Diagnostic System). A diagnostic system DS is defined as the triple DS = ⟨SD, COMPS, OBS⟩, where SD is a propositional theory over a set of variables $V$, COMPS ⊆ $V$, OBS ⊆ $V$, COMPS is the set of assumables, and OBS is the set of observables.

Throughout this paper we will assume that OBS ∩ COMPS = ∅ and SD $\not\models \perp$. Not all propositional theories used as system descriptions are of interest to MBD. Diagnostic systems can be characterized by a restricted set of models, the restriction making the problem of computing diagnosis amenable to algorithms like the one presented in this paper. We consider two main classes of models.

---

1. With formulae in Conjunctive Normal Form (CNF), BCP is implemented through the unit resolution rule.
2. In the MBD literature the assumable variables are also referred to as "component", "failure-mode", or "health" variables. Observable variables are also called "measurable", or "control" variables.

**Definition 2** (Weak-Fault Model). A diagnostic system DS = $\langle$SD, COMPS, OBS$\rangle$ belongs to the class **WFM** iff for COMPS = $\{h_1, h_2, \ldots, h_n\}$, SD is equivalent to $(h_1 \Rightarrow F_1) \wedge (h_2 \Rightarrow F_2) \wedge \ldots \wedge (h_n \Rightarrow F_n)$ and COMPS $\cap V' = \emptyset$, where $V'$ is the set of all variables appearing in the propositional formulae $F_1, F_2, \ldots, F_n$.

Note the conventional selection of the sign of the "health" variables $h_1, h_2, \ldots h_n$. Alternatively, negative literals, e.g., $f_1, f_2, \ldots, f_n$ can be used to express faults, in which case a weak-fault model is in the form $(\neg f_1 \Rightarrow F_1) \wedge \ldots \wedge (\neg f_n \Rightarrow F_n)$. Other authors use "ab" for abnormal or "ok" for healthy.

Weak-fault models are sometimes referred to as models with *ignorance of abnormal behavior* (de Kleer, Mackworth, & Reiter, 1992), or *implicit fault systems*. Alternatively, a model may specify faulty behavior for its components. In the following definition, with the aim of simplifying the formalism throughout this paper, we adopt a slightly restrictive representation of faults, allowing only a single fault-mode per assumable variable. This can be easily generalized by introducing multi-valued logic or suitable encodings (Hoos, 1999).

**Definition 3** (Strong-Fault Model). A diagnostic system DS = $\langle$SD, COMPS, OBS$\rangle$ belongs to the class **SFM** iff SD is equivalent to $(h_1 \Rightarrow F_{1,1}) \wedge (\neg h_1 \Rightarrow F_{1,2}) \wedge \ldots \wedge (h_n \Rightarrow F_{n,1}) \wedge (\neg h_n \Rightarrow F_{n,2})$ such that $1 \leq i, j \leq n, k \in \{1, 2\}$, $\{h_i\} \subseteq$ COMPS, $F_{\{j,k\}}$ is a propositional formula, and none of $h_i$ appears in $F_{j,k}$.

Membership testing for the **WFM** and **SFM** classes can be performed efficiently in many cases, for example, when a model is represented explicitly as in Def. 2 or Def. 3.

## 2.1 A Running Example

We will use the Boolean circuit shown in Fig. 1 as a running example for illustrating all the notions and algorithms in this paper. The subtractor, shown there, consists of seven components: an inverter, two or-gates, two xor-gates, and two and-gates. The expression $h \Rightarrow (o \Leftrightarrow \neg i)$ models the normative (healthy) behavior of an inverter, where the variables $i$, $o$, and $h$ represent input, output and health respectively. Similarly, an and-gate is modeled as $h \Rightarrow [o \Leftrightarrow (i_1 \wedge i_2)]$ and an or-gate by $h \Rightarrow [o \Leftrightarrow (i_1 \vee i_2)]$. Finally, an xor-gate is specified as $h \Rightarrow [o \Leftrightarrow \neg (i_1 \Leftrightarrow i_2)]$.

The above propositional formulae are copied for each gate in Fig. 1 and their variables renamed in such a way as to properly connect the circuit and disambiguate the assumables, thus obtaining a propositional formula for the Boolean subtractor, given by:

$$\begin{aligned}
\text{SD}_w = \{h_1 \Rightarrow [i \Leftrightarrow \neg (y \Leftrightarrow p)]\} \wedge \{h_2 \Rightarrow [d \Leftrightarrow \neg (x \Leftrightarrow i)]\} \wedge [h_3 \Rightarrow (j \Leftrightarrow y \vee p)] \wedge \\
\wedge [h_4 \Rightarrow (m \Leftrightarrow l \wedge j)] \wedge [h_5 \Rightarrow (b \Leftrightarrow m \vee k)] \wedge [h_6 \Rightarrow (x \Leftrightarrow \neg l)] \wedge \\
\wedge [h_7 \Rightarrow (k \Leftrightarrow y \wedge p)]
\end{aligned} \quad (1)$$

A strong-fault model for the Boolean circuit shown in Fig. 1 is constructed by assigning fault-modes to the different gate types. We will assume that, when malfunctioning, the output of an xor-gate has the value of one of its inputs, an or-gate can be stuck-at-one, an and-gate can be stuck-at-zero, and an inverter behaves like a buffer. This gives us the following strong-fault model formula for the Boolean subtractor circuit:

$$\text{SD}_s = \text{SD}_w \wedge [\neg h_1 \Rightarrow (i \Leftrightarrow y)] \wedge [\neg h_2 \Rightarrow (d \Leftrightarrow x)] \wedge (\neg h_3 \Rightarrow j) \wedge \\ \wedge (\neg h_4 \Rightarrow \neg m) \wedge (\neg h_5 \Rightarrow b) \wedge [\neg h_6 \Rightarrow (x \Leftrightarrow l)] \wedge (\neg h_7 \Rightarrow \neg k) \tag{2}$$

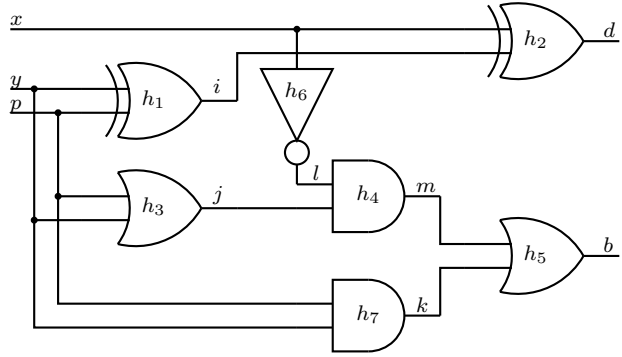

Figure 1: A subtractor circuit

For both models ($\text{SD}_s$ and $\text{SD}_w$), the set of assumable variables is $\text{COMPS} = \{h_1, h_2, \ldots, h_7\}$ and the set of observable variables is $\text{OBS} = \{x, y, p, d, b\}$.

## 2.2 Diagnosis and Minimal Diagnosis

The traditional query in MBD computes terms of assumable variables which are explanations for the system description and an observation.

**Definition 4** (Health Assignment). Given a diagnostic system $\text{DS} = \langle \text{SD}, \text{COMPS}, \text{OBS} \rangle$, an assignment $\omega$ to all variables in COMPS is defined as a health assignment.

A health assignment $\omega$ is a conjunction of propositional literals. In some cases it is convenient to use the set of negative or positive literals in $\omega$. These two sets are denoted as $Lit^-(\omega)$ and $Lit^+(\omega)$, respectively.

In our example, the "all nominal" assignment is $\omega_1 = h_1 \wedge h_2 \wedge \ldots \wedge h_7$. The health assignment $\omega_2 = h_1 \wedge h_2 \wedge h_3 \wedge \neg h_4 \wedge h_5 \wedge h_6 \wedge \neg h_7$ means that the two and-gates from Fig. 1 are malfunctioning. What follows is a formal definition of consistency-based diagnosis.

**Definition 5** (Diagnosis). Given a diagnostic system $\text{DS} = \langle \text{SD}, \text{COMPS}, \text{OBS} \rangle$, an observation $\alpha$, which is an instantiation of some variables in OBS, and a health assignment $\omega$, $\omega$ is a diagnosis iff $\text{SD} \wedge \alpha \wedge \omega \not\models \bot$.

Traditionally, other authors (de Kleer & Williams, 1987) arrive at minimal diagnosis by computing a minimal hitting set of the minimal conflicts (broadly, minimal health assignments incompatible with the system description and the observation), while this paper makes no use of conflicts, hence the equivalent, direct definition above.

There is a total of 96 possible diagnoses given $\text{SD}_w$ and an observation $\alpha_1 = x \wedge y \wedge p \wedge b \wedge \neg d$. Example diagnoses are $\omega_3 = \neg h_1 \wedge h_2 \wedge \ldots \wedge h_7$ and $\omega_4 = h_1 \wedge \neg h_2 \wedge h_3 \wedge \ldots \wedge h_7$. Trivially, given a weak-fault model, the "all faulty" health assignment (in our example $\omega_a = \neg h_1 \wedge \ldots \wedge \neg h_7$) is a diagnosis for any instantiation of the observable variables in OBS (cf. Def. 2).

In the analysis of our algorithm we need the opposite notion of diagnosis, i.e., health assignments inconsistent with a model and an observation. In the MBD literature these assignments are usually called conflicts. Conflicts, however, do not necessarily instantiate all variables in COMPS. As in this paper we always use full health instantiations, the use of the term conflict is avoided to prevent confusion.

In the MBD literature, a range of types of "preferred" diagnosis has been proposed. This turns the MBD problem into an optimization problem. In the following definition we consider the common subset-ordering.

**Definition 6** (Minimal Diagnosis). A diagnosis $\omega^{\subseteq}$ is defined as minimal, if no diagnosis $\tilde{\omega}^{\subseteq}$ exists such that $Lit^-(\tilde{\omega}^{\subseteq}) \subset Lit^-(\omega^{\subseteq})$.

Consider the weak-fault model $SD_w$ of the circuit shown in Fig. 1 and an observation $\alpha_2 = \neg x \wedge y \wedge p \wedge \neg b \wedge d$. In this example, two of the minimal diagnoses are $\omega_5^{\subseteq} = \neg h_1 \wedge h_2 \wedge h_3 \wedge h_4 \wedge \neg h_5 \wedge h_6 \wedge h_7$ and $\omega_6^{\subseteq} = \neg h_1 \wedge h_2 \wedge \ldots \wedge h_5 \wedge \neg h_6 \wedge \neg h_7$. The diagnosis $\omega_7 = \neg h_1 \wedge \neg h_2 \wedge h_3 \wedge h_4 \wedge \neg h_5 \wedge h_6 \wedge h_7$ is non-minimal as the negative literals in $\omega_5^{\subseteq}$ form a subset of the negative literals in $\omega_7$.

Note that the set of all minimal diagnoses characterizes all diagnoses for a weak-fault model, but that does not hold in general for strong-fault models (de Kleer et al., 1992). In the latter case, faulty components may "exonerate" each other, resulting in a health assignment containing a proper superset of the negative literals of another diagnosis not to be a diagnosis. In our example, given $SD_s$ and $\alpha_3 = \neg x \wedge \neg y \wedge \neg p \wedge b \wedge \neg d$, it follows that $\omega_8^{\subseteq} = h_1 \wedge h_2 \wedge \neg h_3 \wedge h_4 \wedge \ldots \wedge h_7$ is a diagnosis, but $\omega_9 = h_1 \wedge h_2 \wedge \neg h_3 \wedge \neg h_4 \wedge \ldots \wedge h_7$ is not a diagnosis, despite the fact that the negative literals in $\omega_9$ form a superset of the negative literals in $\omega_8^{\subseteq}$.

**Definition 7** (Number of Minimal Diagnoses). Let the set $\Omega^{\subseteq}(SD \wedge \alpha)$ contain all minimal diagnoses of a system description SD and an observation $\alpha$. The number of minimal diagnoses, denoted as $|\Omega^{\subseteq}(SD \wedge \alpha)|$, is defined as the cardinality of $\Omega^{\subseteq}(SD \wedge \alpha)$.

Continuing our running example, $|\Omega^{\subseteq}(SD_w \wedge \alpha_2)| = 8$ and $|\Omega^{\subseteq}(SD_s \wedge \alpha_3)| = 2$. The number of non-minimal diagnoses of $SD_w \wedge \alpha_2$ is 61.

**Definition 8** (Cardinality of a Diagnosis). The cardinality of a diagnosis, denoted as $|\omega|$, is defined as the number of negative literals in $\omega$.

Diagnosis cardinality gives us another partial ordering: a diagnosis is defined as *minimal cardinality* iff it minimizes the number of negative literals.

**Definition 9** (Minimal-Cardinality Diagnosis). A diagnosis $\omega^{\leq}$ is defined as minimal-cardinality if no diagnosis $\tilde{\omega}^{\leq}$ exists such that $|\tilde{\omega}^{\leq}| < |\omega^{\leq}|$.

The cardinality of a minimal-cardinality diagnosis computed from a system description SD and an observation $\alpha$ is denoted as $MinCard(SD \wedge \alpha)$. For our example model $SD_w$ and an observation $\alpha_4 = x \wedge y \wedge p \wedge \neg b \wedge \neg d$, it follows that $MinCard(SD_w \wedge \alpha_4) = 2$. Note that in this case all minimal diagnoses are also minimal-cardinality diagnoses.

A minimal cardinality diagnosis is a minimal diagnosis, but the opposite need not hold. In the general case, there are minimal diagnoses which are not minimal-cardinality diagnoses. Consider the example $SD_w$ and $\alpha_2$ given earlier in this section, and the two resulting minimal diagnoses $\omega_5^{\subseteq}$ and $\omega_6^{\subseteq}$. From these two, only $\omega_5^{\subseteq}$ is a minimal-cardinality diagnosis.

**Definition 10** (Number of Minimal-Cardinality Diagnoses). Let the set $\Omega^{\leq}(SD \wedge \alpha)$ contain all minimal-cardinality diagnoses of a system description SD and an observation $\alpha$. The number of minimal-cardinality diagnoses, denoted as $|\Omega^{\leq}(SD \wedge \alpha)|$, is defined as the cardinality of $\Omega^{\leq}(SD \wedge \alpha)$.

Computing the number of minimal-cardinality diagnoses for the running example results in $|\Omega^{\leq}(SD_w \wedge \alpha_2)| = 2$, $|\Omega^{\leq}(SD_s \wedge \alpha_3)| = 2$, and $|\Omega^{\leq}(SD_w \wedge \alpha_4)| = 4$.

## 2.3 Converting Propositional Formulae to Clausal Form

Our approach is related to satisfiability, and SAFARI uses a SAT solver. SAT solvers commonly accept their input in Conjunctive Normal Form (CNF), although there exist SAT solvers that work directly on propositional formulae (Thiffault, Bacchus, & Walsh, 2004). Converting a propositional formula to CNF can be done with (Tseitin, 1983) or without (Forbus & de Kleer, 1993) the introduction of intermediate variables. In both cases important structural information is lost, which may lead to performance degradation when checking if a formula is consistent or when computing a solution.

**Lemma 1.** *A fault-model* $SD = F_1 \wedge F_2 \wedge \ldots \wedge F_n$ *(SD $\in$ **WFM** or SD $\in$ **SFM**) with* $n = |COMPS|$ *component variables can be converted to CNF in time* $O(|COMPS|\zeta)$ *where* $\zeta$ *is the time for converting the largest subformula* $F_i$ *(*$1 \leq i \leq n$*) to CNF.*

*Proof (Sketch).* The conversion of SD to CNF can be done by (1) converting each subformula $F_i$ to CNF and (2) concatenating the resulting CNFs in the final CNF equivalent of SD. The complexity of (1) is $O(n)$ while the complexity of (2) is, in the worst-case, $O(2^m) < \zeta$, where $m$ is the largest number of variables in a subformula $F_i$. As a result, the total time for converting SD is dominated by $\zeta$ and it is linear in $|COMPS|$. $\qquad\square$

Lemma 1 is useful in the cases in which each subformula $F_i$ is small. This is the case in many practical situations where SD is composed of small component models. This is also the case with our experimental benchmark (cf. Sec. 6) where the model of a combinational circuit is the conjunction of fault models of simple logic gates ($x$-bit and-gates, typically $x < 10$, xor-gates, etc.). Ideally, SAFARI would use a non-CNF SAT solver, but for practical reasons we have constrained our reasoning to diagnostic models with concise CNF encodings.

Consider, for example, the formula $(x_1 \wedge y_1) \vee (x_2 \wedge y_2) \vee \cdots \vee (x_n \wedge y_n)$, which is in Disjunctive Normal Form[3] (DNF) and, converted to CNF, has $2^n$ clauses. Although similar examples of propositional formulae having exponentially many clauses in their CNF representations are easy to find, they are artificial and are rarely encountered in MBD. Furthermore, the Boolean circuits with which we have tested the performance of SAFARI do not show exponential blow-up when converted to CNF.

---

3. Note that all DNF formulae are also propositional formulae.

### 2.4 Complexity of Diagnostic Inference

This section discusses the complexity of the problems in which we are interested, namely the problem of computing a single or the set of all minimal diagnoses, using two minimality criteria, subset-minimality ($\subseteq$) and cardinality-minimality ($\leq$). We assume as input a CNF formula defined over a variable set $V$, of which $\gamma = |\text{COMPS}|$ are assumable (or fault) variables. Table 1 introduces the notation that we use to define these 4 types of diagnosis.

Table 1: Summary of definitions of types of diagnosis of interest

| Symbol | Diagnoses | Preference Criterion |
|--------|-----------|----------------------|
| $\omega^{\subseteq}$ | 1 | $\subseteq$ (subset-minimality) |
| $\omega^{\leq}$ | 1 | $\leq$ (cardinality-minimality) |
| $\Omega^{\subseteq}$ | all | $\subseteq$ (subset-minimality) |
| $\Omega^{\leq}$ | all | $\leq$ (cardinality-minimality) |

The complexity of computing the set of all diagnoses is harder than computing a single diagnosis, since the number of diagnoses is, in the worst case, exponential in the input size (number of components). This problem is bounded from below by the problem of counting the number of diagnoses. This problem has been shown to be #co-$NP$-Complete (Hermann & Pichler, 2007).

If we restrict our clauses to be Horn or definite Horn, then we can reduce the complexity of the problems that we are solving, at the expense of decreased model expressiveness. Under a Horn-clause restriction, for SD $\in$ **WFM**, determining if a first minimal diagnosis exists is in $P$. Under the same restriction, for SD $\in$ **SFM**, deciding if a first minimal diagnosis exists is $NP$-hard (Friedrich et al., 1990). In both cases (SD $\in$ **WFM**, **SFM**) deciding if a next diagnosis exists is $NP$-hard.

The diagnosis problems of interest in this article are intractable in the worst-case. The complexity of a closely-related problem, Propositional Abduction Problems (PAPs), has been studied by Eiter and Gottlob (1995). They show that for a propositional PAP, the problem of determining if a solution exists is $\Sigma_2^P$-complete. Computing a minimal diagnosis is a search problem, and hence it is more difficult to pose a decision question for proving complexity results. Consequently, one can just note that computing a diagnosis minimal with respect to $\subseteq$ / $\leq$ requires $O(\log |\text{COMPS}|)$ calls to an $NP$ oracle (Eiter & Gottlob, 1995), asking the oracle at each step if a diagnosis containing at most $k$ faulty components exists.

Results on abduction problems indicate that the task of approximate diagnosis is intractable. Roth (1996) has addressed the problems of abductive inference, and of approximating such inference. Roth focuses on counting the number of satisfying assignments for a range of AI problems, including some instances of PAPs. In addition, Roth shows that approximating the number of satisfying assignments for these problems is intractable.

Abdelbar (2004) has studied the complexity of approximating Horn abduction problems, showing that even for a particular Horn restriction of the propositional problem of interest, the approximation problem is intractable. In particular, for an abduction problem with costs assigned to the assumables (which can be used to model the preference-ordering $\leq$), he has examined the complexity of finding the Least Cost Proof (LCP) for the evidence (OBS), where the cost of a proof is taken to be the sum of the costs of all hypotheses that must be assumed in order to complete the proof. For this problem he has shown that it is *NP*-hard to approximate an LCP within a fixed ratio $r$ of the cost of an optimal solution, for any $r < 0$.

SAFARI approximates the intractable problems denoted in Table 1. We show that for **WFM**, SAFARI can efficiently compute a single diagnosis that is minimal under $\subseteq$ by using a satisfiability oracle. For SD $\in$ **SFM**, SAFARI generates a sound but possibly sub-optimal diagnosis (or set of diagnoses). We have referred to papers indicating that it is intractable to approximate, within a fixed ratio, a minimal diagnosis. In the following, we adopt a stochastic approach that cannot provide fixed-ratio guarantees. However, SAFARI trades off optimality for efficiency and can compute most diagnoses with high likelihood.

## 3. Stochastic MBD Algorithm

In this section we discuss an algorithm for computing multiple-fault diagnoses using stochastic search.

### 3.1 A Simple Example (Continued)

Consider the Boolean subtractor shown in Fig. 1, its weak-fault model $SD_w$ given by (1), and the observation $\alpha_4$ from the preceding section. The four minimal diagnoses associated to $SD_w \wedge \alpha_4$ are: $\omega_1 = \neg h_1 \wedge h_2 \wedge h_3 \wedge h_4 \wedge \neg h_5 \wedge h_6 \wedge h_7$, $\omega_2 = h_1 \wedge \neg h_2 \wedge h_3 \wedge h_4 \wedge \neg h_5 \wedge h_6 \wedge h_7$, $\omega_3 = \neg h_1 \wedge h_2 \wedge \ldots \wedge h_6 \wedge \neg h_7$, and $\omega_4 = h_1 \wedge \neg h_2 \wedge h_3 \wedge \ldots \wedge h_6 \wedge \neg h_7$.

A naïve deterministic algorithm would check the consistency of all the $2^{|COMPS|}$ possible health assignments for a diagnostic problem, 128 in the case of our running example. Furthermore, most deterministic algorithms first enumerate health assignments of small cardinality but with high a priori probability, which renders these algorithms impractical in situations when the minimal diagnosis is of a higher cardinality. Such performance is not surprising even when using state-of-the art MBD algorithms which utilize, for example conflict learning, or partial compilation, considering the bad worst-case complexity of finding all minimal diagnoses (cf. Sec. 2.4).

In what follows, we will show a two-step diagnostic process that requires fewer consistency checks. The first step involves finding a random non-minimal diagnosis as a starting point (cf. Sec. 3.2 for details on computing random SAT solutions with equal likelihood). The second step attempts to minimize the fault cardinality of this diagnosis by repeated modification of the diagnosis.

The first step is to find one random, possibly non-minimal diagnosis of $SD_w \wedge \alpha_4$. Such a diagnosis we can obtain from a classical DPLL solver after modifying it in two ways: (1) not only determine if the instance is satisfiable but also extract the satisfying solution and (2) find a *random* satisfiable solution every time the solver is invoked. Both modifications are trivial, as DPLL solvers typically store their current variable assignments and it is easy to choose a variable and value randomly (according to a uniform distribution) instead of deterministically when branching. The latter modification may possibly harm a DPLL variable or value selection heuristics, but later in this paper we will see that this is of no concern for the type of problems we are considering as diagnostic systems are typically underconstrained.

In the subtractor example we call the DPLL solver with $SD_w \wedge \alpha_4$ as an input and we consider the random solution (and obviously a diagnosis) $\omega_5 = \neg h_1 \wedge h_2 \wedge \neg h_3 \wedge h_4 \wedge h_5 \wedge \neg h_6 \wedge \neg h_7$ ($|\omega_5| = 4$). In the second step of our stochastic algorithm, we will try to minimize $\omega_5$ by repetitively choosing a random negative literal, "flipping" its value to positive (thus obtaining a candidate with a smaller number of faults), and calling the DPLL solver. If the new candidate is a diagnosis, we will try to improve further this newly discovered diagnosis, otherwise we will mark the attempt a "failure" and choose another negative literal. After some constant number of "failures" (two in this example), we will terminate the search and will store the best diagnosis discovered so far in the process.

After changing the sign of $\neg h_7$ in $\omega_5$ we discover that the new health assignment is not consistent with $SD_w \wedge \alpha_4$, hence it is not a diagnosis and we discard it. Instead, the algorithm attempts changing $\neg h_6$ to $h_6$ in $\omega_5$, this time successfully obtaining a new diagnosis $\omega_6 = \neg h_1 \wedge h_2 \wedge \neg h_3 \wedge h_4 \wedge h_5 \wedge h_6 \wedge \neg h_7$ of cardinality 3. Next the algorithm tries to find a diagnosis of even smaller cardinality by randomly choosing $\neg h_1$ and $\neg h_7$ in $\omega_6$, respectively, and trying to change their sign, but both attempts return an inconsistency. Hence the "climb" is aborted and $\omega_6$ is stored as the current best diagnosis.

Repeating the process from another random initial DPLL solution, gives us a new diagnosis $\omega_7 = \neg h_1 \wedge \neg h_2 \wedge h_3 \wedge \neg h_4 \wedge h_5 \wedge h_6 \wedge \neg h_7$. Changing the sign of $\neg h_7$, again, leads to inconsistency, but the next two "flips" (of $\neg h_4$ and $\neg h_2$) lead to a double-fault diagnosis $\omega_8 = \neg h_1 \wedge h_2 \wedge \ldots \wedge h_6 \wedge \neg h_7$. The diagnosis $\omega_8$ can not be improved any further as it is minimal. Hence the next two attempts to improve $\omega_8$ fail and $\omega_8$ is stored in the result.

This process is illustrated in Fig. 2, the search for $\omega_6$ is on the left and for $\omega_8$ on the right. Gates which are shown in solid black are "suspected" as faulty when the health assignment they participate in is tested for consistency, and inconsistent candidates are crossed-out. Let us consider the result. We have found two diagnoses: $\omega_6$ and $\omega_8$, where $\omega_6$ is not a minimal diagnosis. This we have done at the price of 11 calls to a DPLL subroutine. The suboptimal diagnosis $\omega_6$ is of value as its cardinality is near the one of a minimal diagnosis. Hence we have demonstrated a way to find an approximation of all minimal diagnoses, while drastically reducing the number of consistency checks in comparison to a deterministic algorithm, sacrificing optimality. Next we will formalize our experience into an algorithm, the behavior of which we will analyze extensively in the section that follows.

Diagnosing a strong-fault model is known to be strictly more difficult than a weak-fault model (Friedrich et al., 1990). In many diagnostic instances this problem is alleviated by the fact that there exist, although without a guarantee, continuities in the diagnostic search space similar to the one in the weak-fault models. Let us discuss the process of finding a minimal diagnosis of the subtractor's strong-fault model $SD_s$ and the observation $\alpha_2$ (both from Sec. 2.1).

The six distinct diagnoses $\omega_9, \ldots, \omega_{14}$ of $SD_s$ and $\alpha_2$ are shown in Fig. 3. Of these only $\omega_9$ and $\omega_{10}$ are minimal such that $|\omega_9| = |\omega_{10}| = 3$. It is visible in Fig. 3 that in all diagnoses component variables $h_2$ and $h_5$ are false, while $h_1$ and $h_7$ are true (healthy). Hence, any satisfying assignment of $SD_s \wedge \alpha_2$ would contain $h_1 \wedge \neg h_2 \wedge \neg h_5 \wedge h_7$. Starting from the maximal-cardinality diagnosis $\omega_{14}$, we must "flip" the variables $h_3$, $h_4$, and $h_6$ in order to reach the two minimal diagnoses. The key insight is that, as shown in Fig. 3, this is always

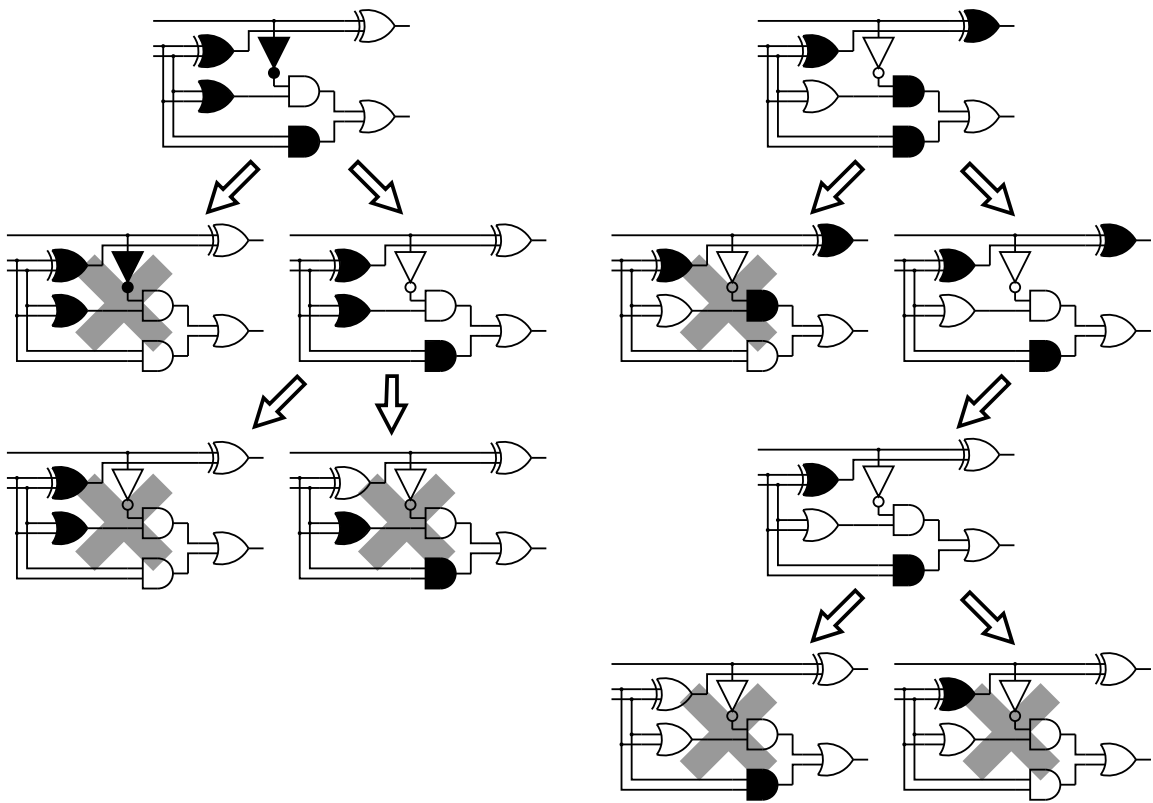

Figure 2: An example of a stochastic diagnostic process

|  | $h_2$ | $h_5$ | $h_4$ | $h_6$ | $h_3$ | $h_1$ | $h_7$ |  |  | $h_2$ | $h_5$ | $h_4$ | $h_6$ | $h_3$ | $h_1$ | $h_7$ |
|---|---|---|---|---|---|---|---|---|---|---|---|---|---|---|---|---|
| $\omega_9$ | ✗ | ✗ | ✗ | ✓ | ✗ | ✓ | ✓ |  | $\omega_{10}$ | ✗ | ✗ | ✓ | ✗ | ✗ | ✓ | ✓ |
| $\omega_{13}$ | ✗ | ✗ | ✓ | ✓ | ✗ | ✓ | ✓ |  | $\omega_{12}$ | ✗ | ✗ | ✓ | ✗ | ✓ | ✓ | ✓ |
| $\omega_{14}$ | ✗ | ✗ | ✓ | ✓ | ✓ | ✓ | ✓ |  | $\omega_{14}$ | ✗ | ✗ | ✓ | ✓ | ✓ | ✓ | ✓ |

|  | $h_2$ | $h_5$ | $h_4$ | $h_6$ | $h_3$ | $h_1$ | $h_7$ |  |  | $h_2$ | $h_5$ | $h_4$ | $h_6$ | $h_3$ | $h_1$ | $h_7$ |
|---|---|---|---|---|---|---|---|---|---|---|---|---|---|---|---|---|
| $\omega_9$ | ✗ | ✗ | ✗ | ✓ | ✗ | ✓ | ✓ |  | $\omega_{10}$ | ✗ | ✗ | ✓ | ✗ | ✗ | ✓ | ✓ |
| $\omega_{11}$ | ✗ | ✗ | ✗ | ✓ | ✓ | ✓ | ✓ |  | $\omega_{13}$ | ✗ | ✗ | ✓ | ✓ | ✗ | ✓ | ✓ |
| $\omega_{14}$ | ✗ | ✗ | ✓ | ✓ | ✓ | ✓ | ✓ |  | $\omega_{14}$ | ✗ | ✗ | ✓ | ✓ | ✓ | ✓ | ✓ |

Figure 3: Diagnoses of a strong-fault model

possible by "flipping" a single literal at a time from health to faulty and receiving another
consistent assignment (diagnosis).

In what follows we will formalize our experience so far in a stochastic algorithm for
finding minimal diagnoses.

## 3.2 A Greedy Stochastic Algorithm

Algorithm 1 shows the pseudocode of SAFARI.

---

**Algorithm 1** SAFARI: A greedy stochastic hill climbing algorithm for approximating the set of minimal diagnoses

---

1: **function** SAFARI(DS, $\alpha$, $M$, $N$) **returns** a trie
  **inputs:** DS = $\langle$SD, COMPS, OBS$\rangle$, diagnostic system
    $\alpha$, term, observation
    $M$, integer, climb restart limit
    $N$, integer, number of tries
  **local variables:** $\text{SD}_{\text{cnf}}$, CNF
       $m, n$, integers
       $\omega, \omega'$, terms
       $R$, set of terms, result
2:  $\text{SD}_{\text{cnf}} \leftarrow$ WFFTOCNF(SD)
3:  **for** $n = 1, 2, \ldots, N$ **do**
4:   $\omega \leftarrow$ RANDOMDIAGNOSIS($\text{SD}_{\text{cnf}}, \alpha$)     ▷ Get a random SAT solution.
5:   $m \leftarrow 0$
6:   **while** $m < M$ **do**
7:    $\omega' \leftarrow$ IMPROVEDIAGNOSIS($\omega$)   ▷ Flip an "unflipped" health variable.
8:    **if** $\text{SD}_{\text{cnf}} \wedge \alpha \wedge \omega' \not\models \bot$ **then**    ▷ Consistency check.
9:     $\omega \leftarrow \omega'$
10:     $m \leftarrow 0$
11:    **else**
12:     $m \leftarrow m + 1$
13:    **end if**
14:   **end while**
15:   **unless** ISSUBSUMED($R, \omega$) **then**
16:    ADDTOTRIE($R, \omega$)
17:    REMOVESUBSUMED($R, \omega$)
18:   **end unless**
19:  **end for**
20:  **return** $R$
21: **end function**

---

SAFARI accepts two input parameters: $M$ and $N$. There are $N$ independent searches that start from randomly generated starting points. The algorithm tries to improve the cardinality of the initial diagnoses (while preserving their consistency) by randomly "flipping" fault literals. The change of a sign of literal is done in one direction only: from faulty to healthy. Each attempt to find a minimal diagnosis terminates after $M$ unsuccessful attempts to "improve" the current diagnosis stored in $\omega$. Thus, increasing $M$ will lead to a better exploration of the search space and, possibly, to diagnoses of lower cardinality, while decreasing it will improve the overall speed of the algorithm.

SAFARI uses a number of utility functions. WFFTOCNF converts the propositional formula in SD to CNF (cf. Sec 2.3). The IMPROVEDIAGNOSIS subroutine takes a term $\omega$ as an argument and changes the sign of a random negative literal in $\omega$. If there are no negative literals, the function returns its original argument.

The implementation of RANDOMDIAGNOSIS uses a modified DPLL solver returning a random SAT solution of SD ∧ α. Consider the original DPLL algorithm (Davis, Logemann, & Loveland, 1962) without the unit resolution rule. One can show that if, in the event of branching, the algorithm chooses unassigned variables and their polarity with equal probability, the DPLL algorithm is equally likely to compute any satisfiable solution (if such exists). Note that the order in which variables are assigned does not matter. Of course, the DPLL algorithm may end-up with a partial assignment, i.e., some of the variables are "don't care". This is not a problem because the partial assignment can be extended to a full satisfiable assignment by randomly choosing the signs of the unassigned variables from a uniform distribution. Taking into consideration the unit resolution rule, does not change the likelihood of the modified DPLL solver finding a particular solution because it only changes the order in which variables are assigned. A formal proof that this modified DPLL solver computes a SAT assignment with equal probability is beyond the scope of this paper, but the idea is to build a probabilistic model of the progress of the DPLL solver. This probabilistic model is a balanced tree where nodes iterate between branching and performing unit resolution (assigning values to zero or more unit clauses). As the branching probability is set to be equal and all leaf nodes (SAT solutions) are at equal depth, one can show the equal likelihood of arriving to any SAT solution. As most up-to-date SAT solvers are based on DPLL, creating a randomized DPLL solver that computes any satisfiable solution with equal probability is not difficult. Of course, random polarity decisions may effect negatively branching heuristics (Marques-Silva, 1999) but such analysis is also beyond the scope of this paper.

Similar to deterministic methods for MBD, SAFARI uses a SAT-based procedure for checking the consistency of SD ∧ α ∧ ω. To increase the implementation efficiency of SAFARI, we combine a BCP-based LTMS engine (McAllester, 1990) and a full-fledged DPLL solver in two-stage consistency checking. Experimentation shows that combining LTMS and DPLL in such a way allows an order-of-magnitude SAFARI speed-up compared to pure DPLL, while the soundness and completeness properties of consistency checking are preserved.

We have implemented the two-stage consistency checking as follows. First, SAFARI calls a BCP-based LTMS (Forbus & de Kleer, 1993) to check if SD ∧ α ∧ ω ⊨⊥. If the result is UNSAT then the candidate ω is not a diagnosis.[4] If the LTMS result is not UNSAT, it means that the consistency of the candidate is unknown and a call to a complete DPLL engine is needed. For the full DPLL checking we use POSIT (Freeman, 1995) or MINISAT (Eén & Sörensson, 2003).

SAFARI benefits from the two-stage SAT procedure because a typical MBD instance involves many consistency checks ($O(|\text{COMPS}|^2)$ for $N = 1, M = |\text{COMPS}|$). As SD ∧ α does not change during the search and each time only a small number of assumption clauses have to be updated, the incremental nature of LTMS greatly improves the search efficiency. Even though the DPLL running time per instance is the same as LTMS (DPLL performs BCP when doing unit propagation), DPLL construction is expensive and should be avoided when possible. DPLL initialization is typically slow as it involves building data structures for clauses and variables, counting literals, initializing conflict databases, etc. On the other hand, our implementation of LTMS is both incremental (does not have to be reinitialized

---

4. It can be shown that if a BCP consistency check of SD ∧ α ∧ ω returns UNSAT, then the formula is UNSAT (the opposite is not necessarily true).

before each consistency check) and efficient as it maintains only counters for each clause. Each counter keeps the number of unassigned literals. Assigning a value to a variable requires decrementing some or all of the clause counters. If a counter becomes zero, a contradiction handler is signaled.

There is no guarantee that two diagnostic searches, starting from random diagnoses, would not lead to the same minimal diagnosis. To prevent this, we store the generated diagnoses in a trie $R$ (Forbus & de Kleer, 1993), from which it is straightforward to extract the resulting diagnoses by recursively visiting its nodes. A diagnosis $\omega$ is added to the trie $R$ by the function AddToTrie, iff no subsuming diagnosis is contained in $R$ (the IsSubsumed subroutine checks on that condition). After adding a diagnosis $\omega$ to the resulting trie $R$, all diagnoses contained in $R$ and subsumed by $\omega$ are removed by a call to RemoveSubsumed.

### 3.3 Basic Properties of the Greedy Stochastic Search

Before we continue with the topics of completeness and optimality, we show that Safari is sound, i.e., it returns diagnoses only.

**Lemma 2** (Soundness)**.** Safari *is sound.*

*Proof (Sketch).* The consistency check in line 8 of Alg. 1 guarantees that only terms $\omega$ for which it holds that $SD \wedge \alpha \wedge \omega \not\models \bot$ will be added to the result set $R$. According to Def. 5 these terms $\omega$ are diagnoses. $\qquad\square$

One of the key factors in the success of the proposed algorithm is the exploitation of the continuity of the search-space of diagnosis models, where by continuity we mean that we can monotonically reduce the cardinality of a non-minimal diagnosis. Through the exploitation of this continuity property, Safari can be configured to guarantee finding a minimal diagnosis in weak fault models in a polynomial number of calls to a satisfiability oracle.

The hypothesis which comes next is well studied in prior work (de Kleer et al., 1992), as it determines the conditions under which minimal diagnoses represent all diagnoses of a model and an observation. This paper is interested in the hypothesis from the computational viewpoint: it defines a class of models for which it is possible to establish a theoretical bound on the optimality and performance of Safari.

**Hypothesis 1** (Minimal Diagnosis Hypothesis)**.** Let $DS = \langle SD, COMPS, OBS \rangle$ be a diagnostic system and $\omega'$ a diagnosis for an arbitrary observation $\alpha$. The Minimal Diagnosis Hypothesis (MDH) holds in DS iff for any health assignment $\omega$ such that $Lit^-(\omega) \supset Lit^-(\omega')$, $\omega$ is also a diagnosis.

It is easy to show that MDH holds for all weak-fault models. There are other theories $SD \notin \mathbf{WFM}$ for which MDH holds (e.g., one can directly construct a theory as a conjunction of terms for which MDH holds). Unfortunately, no necessary condition is known for MDH to hold in an arbitrary SD. The lemma which comes next is a direct consequence of MDH and weak-fault models.

**Lemma 3.** *Given a diagnostic system* $DS = \langle SD, COMPS, OBS \rangle$, $SD \in \mathbf{WFM}$, *and a diagnosis* $\omega$ *for some observation* $\alpha$, *it follows that* $\omega$ *is non-minimal iff another diagnosis* $\omega'$ *can be obtained by changing the sign of exactly one negative literal in* $\omega$.

*Proof (Sketch).* From Def. 2 and SD $\in$ **WFM**, it follows that if $\omega$ is a minimal diagnosis, any diagnosis $\omega'$ obtained by flipping one positive literal in $\omega$ is also a diagnosis. Applying the argument in the other direction gives us the above statement. $\qquad\square$

SAFARI operates by performing subset flips on non-minimal diagnoses, attempting to compute minimal diagnoses. We next formalize this notion of flips, in order to characterize when SAFARI will be able to compute a minimal diagnosis.

**Definition 11** (Subset Flip $\Phi_\Downarrow$). Given a diagnostic system DS $= \langle$ SD, COMPS, OBS $\rangle$ and a health assignment $\omega$ with a non-empty set of negative literals ($Lit^-(\omega) \neq \emptyset$), a subset flip $\Phi_\Downarrow$ turns one of the negative literals in $\omega$ to a positive literal, i.e., it creates a health assignment $\omega'$ with one more positive literal.

We next characterize flips based on whether they produce consistent models after the flip.

**Definition 12** (Valid Subset Flip). Given a diagnostic system DS $= \langle$ SD, COMPS, OBS $\rangle$, an observation $\alpha$, and a non-minimal diagnosis $\omega$, a valid flip exists if we can perform a subset flip in $\omega$ to create $\omega'$ such that SD $\wedge \alpha \wedge \omega' \not\models \perp$.

Given these notions, we can define continuity of the diagnosis search space in terms of literal flipping.

**Definition 13** (Continuity). A system model SD and an observation $\alpha$ satisfy the continuity property with respect to the set of diagnoses $\Omega^{\subseteq}(\text{SD} \wedge \alpha)$, iff for any diagnosis $\omega_k \in \Omega(\text{SD} \wedge \alpha)$ there exists a sequence $\Phi = \langle \omega_1, \omega_2, \cdots, \omega_{k-1}, \omega_k, \omega_{k+1}, \cdots, \omega_n \rangle$, such that for $i = 1, 2, \cdots, n-1$, it is possible to go from $\omega_i$ to $\omega_{i+1}$ via a valid subset flip, $\omega_i \in \Omega(\text{SD} \wedge \alpha)$, and $\omega_n \in \Omega^{\subseteq}(\text{SD} \wedge \alpha)$.

The above definition allows for trivial continuity in the cases when a model and an observation lead to minimal diagnoses only (no non-minimal diagnoses). As we will see in Sec. 6, models and observations such that all diagnoses are minimal are rare in practice (of course, such problems can be created artificially). Note that the SAFARI algorithm still works and its theoretical properties are preserved even in the case of trivial continuity.

Given Def. 13, we can easily show the following two lemmata:

**Lemma 4.** *If* SD *satisfies MDH, then it satisfies the continuity property.*

*Proof.* Follows directly from Hypothesis 1 and Def 13. $\qquad\blacksquare$

**Lemma 5.** SD $\in$ **WFM** *satisfies the continuity property.*

*Proof (Sketch).* It is straightforward to show that if SD $\in$ **WFM** then SD satisfies MDH. Then from Lemma 4 it follows that SD satisfies the continuous property. $\qquad\blacksquare$

Our greedy algorithm starts with an initial diagnosis and then randomly flips faulty assumable variables. We now use the MDH property to show that, starting with a non-minimal diagnosis $\omega$, the greedy stochastic diagnosis algorithm can monotonically reduce the size of the "seed" diagnosis to obtain a minimal diagnosis through appropriately flipping a fault variable from faulty to healthy; if we view this flipping as search, then this search is continuous in the diagnosis space.

**Proposition 1.** *Given a diagnostic system* DS $= \langle$SD, COMPS, OBS$\rangle$, *an observation* $\alpha$, *and* SD $\in$ **WFM**, SAFARI *configured with* $M = |$COMPS$|$ *and* $N = 1$ *returns one minimal diagnosis.*

*Proof.* The diagnosis improvement loop starts, in the worst case, from a health assignment $\omega$ which is a conjunction of negative literals only. Necessarily, in this case, $\omega$ is a diagnosis as SD $\in$ **WFM**. A diagnosis $\omega'$ that is subsumed by $\omega$ would be found with at most $M$ consistency checks (provided that $\omega'$ exists) as $M$ is set to be equal to the number of literals in $\omega$ and there are no repetitions in randomly choosing of which literal to flip next. If, after trying all the negative literals in $\omega$, there is no diagnosis, then from Lemma 3 it follows that $\omega$ is a minimal diagnosis.

Through a simple inductive argument, we can continue this process until we obtain a minimal diagnosis. □

From Proposition 1 it follows that there is an upper bound of $O(|$COMPS$|)$ consistency checks for finding a single minimal diagnosis. In most of the practical cases, however, we are interested in finding an approximation to *all* minimal-cardinality diagnoses. As a result the complexity of the optimally configured SAFARI algorithm becomes $O(|$COMPS$|^2 S)$, where $S$ is the number of minimal-cardinality diagnoses for the given observation. Section 5 discusses in more detail the computation of multiple minimal-cardinality diagnoses.

The number of assumable variables in a system of practical significance may exceed thousands, rendering an optimally configured SAFARI computationally too expensive. In Sec 4 we will see that while it is more computationally efficient to configure $M < |$COMPS$|$, it is still possible to find a minimal diagnosis with high probability.

It is simple to show that flip-based search algorithms are complete for continuous diagnosis search spaces given weak fault models, i.e., SD $\in$ **WFM**, and models that follow MDH, i.e., Lemma 3. We can formally characterize the guarantee of finding a minimal diagnosis with SAFARI in terms of a continuous diagnosis space. Note that this is a sufficient, but not necessary, condition; for example, we may configure SAFARI to flip multiple literals at a time to circumvent problems of getting trapped in discontinuous diagnosis spaces.

**Theorem 1.** *Given a diagnostic system* DS $= \langle$SD, COMPS, OBS$\rangle$, *and a starting diagnosis* $\omega$, SAFARI *configured with* $M = |$COMPS$|$ *and* $N = 1$ *is guaranteed to compute a minimal diagnosis if the diagnosis space is continuous.*

*Proof.* Given an initial diagnosis $\omega$, SAFARI attempts to compute a minimal diagnosis by performing subset flips. If the diagnosis space is continuous, then we know that there exists a sequence of valid flips leading to a minimal diagnosis. Hence SAFARI is guaranteed to find a minimal diagnosis from $\omega$. □

Finally, we show that SAFARI provides a strong probabilistic guarantee of computing all minimal diagnoses.

**Theorem 2.** *The probability of* SAFARI, *configured with* $M = |$COMPS$|$, *of computing all minimal diagnoses of a diagnostic system* DS $= \langle$SD, COMPS, OBS$\rangle$ *and an observation* $\alpha$ *is denoted as* Pr$^\star$. *Given a continuous diagnosis space* $\Omega($SD$, \alpha)$, *it holds that* Pr$^\star \to 1$ *for* $N \to \infty$.

*Proof (Sketch).* Since (1) the search space is continuous, (2) at each step there is a non-zero probability of flipping any unflipped literal, and (3) there is a polynomial upper bound of steps (|COMPS|) for computing a diagnosis, SAFARI can compute any non-minimal diagnosis with non-zero probability. Hence as $N \rightarrow \infty$, SAFARI will compute all minimal diagnoses. $\qquad \blacksquare$

### 3.4 Complexity of Inference Using Greedy Stochastic Search

We next look at the complexity of SAFARI, and its stochastic approach to computing sound but incomplete diagnoses. We show that the primary determinant of the inference complexity is the consistency checking. SAFARI randomly computes a partial assignment $\pi$, and then checks if $\pi$ can be extended to create a satisfying assignment during each consistency check, i.e., it checks the consistency of $\pi$ with SD. This is solving the satisfiability problem (SAT), which is NP-complete (Cook, 1971). We will show how we can use incomplete satisfiability checking to reduce this complexity, at the cost of completeness guarantees.

In the following, we call $\Theta$ the complexity of a consistency check, and assume that there are $\gamma$ components that can fail, i.e., $\gamma = $ |COMPS|.

**Lemma 6.** *Given a diagnostic system* DS $= \langle$SD, COMPS, OBS$\rangle$ *with* SD $\in$ **WFM***, the worst-case complexity of finding any minimal diagnosis is* $O(\gamma^2 \Theta)$, *where* $\Theta$ *is the cost of a consistency check.*

*Proof.* There is an upper bound of $\gamma$ succeeding consistency checks for finding a single minimal diagnosis since there is a maximum of $\gamma$ steps for computing the "all healthy" diagnosis. As SAFARI performs a consistency check after each flip and at each step the algorithm must flip at most $\gamma$ literals, the total complexity is $O(\gamma^2 \Theta)$. $\qquad \blacksquare$

In most practical cases, however, we are interested in finding an approximation to *all* minimal-cardinality diagnoses. As a result the complexity of the optimally configured SAFARI algorithm becomes $O\left(\gamma\binom{|\omega|}{\gamma}\Theta\right)$, where $|\omega|$ is the cardinality of the minimal-cardinality diagnoses for the given observation (cf. Sec. 6.6).

The complexity of BCP is well-known, allowing us to get more precise bounds on the worst-case complexity of computing one minimal-diagnosis with SAFARI. In what follows we will assume that SD is represented in CNF (cf. Sec. 2.3).

**Lemma 7.** *Given a diagnostic system* DS $= \langle$SD, COMPS, OBS$\rangle$, SD $\in$ **WFM***, and* SD *having $c$ clauses and $n$ variables, the worst-case complexity under* **WFM** *of finding any minimal diagnosis is* $O(\gamma^2 cn)$ *when using BCP for consistency checks.*[5]

*Proof (Sketch).* An implementation of BCP (Forbus & de Kleer, 1993) maintains a total of $c$ counters for the number of unsatisfied literals in each clause. A consistency check requires decrementing some or all counters for each of the $n$ variables in SD. This gives us an upper bound of $O(cn)$ on the execution time of BCP. Combining the complexity of BCP with Lemma 6 gives us the desired result. $\qquad \blacksquare$

---

5. More efficient implementations of BCP exist (Zhang & Stickel, 1996).

## 4. Optimality Analysis (Single Diagnosis)

In contrast to deterministic algorithms, in the SAFARI algorithm there is no absolute guarantee that the optimum solution (minimal diagnosis) is found. Below we will provide an intuition behind the performance of the SAFARI algorithm by means of an approximate, analytical model that estimates the probability of reaching a diagnostic solution of specific minimality.

### 4.1 Optimality of Safari in Weak-Fault Models

We will start by considering a single run of the algorithm without retries where we will assume the existence of only one minimal diagnosis. Next, we will extend the model by considering retries.

#### 4.1.1 Basic Model

Consider a diagnostic system $DS = \langle SD, COMPS, OBS \rangle$ such that $SD \in \mathbf{WFM}$, and an observation $\alpha$ such that $\alpha$ manifests only one minimal diagnosis $\omega$. For the argument that follows we will configure SAFARI with $M = 1$, $N = 1$, and we will assume that the starting solution is the trivial "all faulty" diagnosis.

When SAFARI randomly chooses a faulty variable and flips it, we will be saying that it is a "success" if the new candidate is a diagnosis, and a "failure" otherwise. Let $k$ denote the number of steps that the algorithm successfully traverses in the direction of the minimal diagnosis of cardinality $|\omega|$. Thus $k$ also measures the number of variables whose values are flipped from faulty to healthy in the process of climbing.

Let $f(k)$ denote the probability distribution function (pdf) of $k$. In the following we derive the probability $p(k)$ of successfully making a transition from $k$ to $k + 1$. A diagnosis at step $k$ has $k$ positive literals and $|COMPS| - k$ negative literals. The probability of the next variable flip being successful equals the probability that the next negative to positive flip out of the $H - k$ negative literals does not conflict with a negative literal belonging to a diagnosis solution $\omega$. Consequently, of the $|\omega| - k$ literals only $COMPS| - |\omega| - k$ literals are allowed to flip, and therefore the success probability equals:

$$p(k) = \frac{|COMPS| - |\omega| - k}{|COMPS| - k} = 1 - \frac{|\omega|}{|COMPS| - k} \tag{3}$$

The search process can be modeled in terms of the Markov chain depicted in Fig. 4, where $k$ equals the state of the algorithm. Running into an inconsistency is modeled by the transitions to the state denoted "fail".

The probability of exactly attaining step $k$ (and subsequently failing) is given by:

$$f(k) = (1 - p(k + 1)) \prod_{i=0}^{k} p(i) \tag{4}$$

Substituting (3) in (4) gives us the pdf of $k$:

$$f(k) = \frac{|\omega|}{|COMPS| - k + 1} \prod_{i=0}^{k} \left[ 1 - \frac{|\omega|}{|COMPS| - i} \right] \tag{5}$$

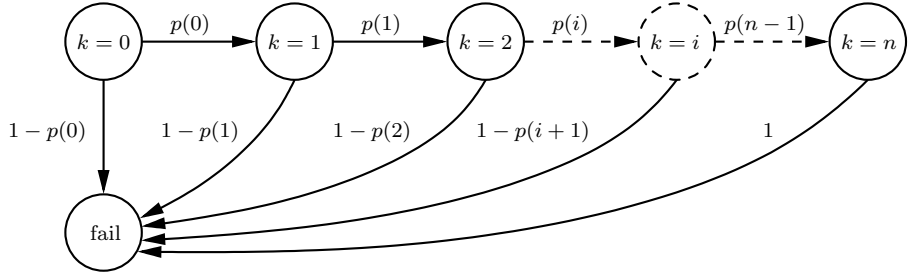

Figure 4: Model of a Safari run for $M = 1$ and a single diagnosis $\omega$ ($n = |\text{COMPS}| - |\omega|$)

At the optimum goal state $k = |\text{COMPS}| - |\omega|$ the failure probability term in (5) is correct as it equals unity.

If $p$ were independent of $k$, $f$ would be geometrically distributed, which implies that the chance of reaching a goal state $k = |\text{COMPS}| - |\omega|$ is slim. However, the fact that $p$ decreases with $k$ moves the probability mass to the tail of the distribution, which works in favor of reaching higher-$k$ solutions. For instance, for single-fault solutions ($|\omega| = 1$) the distribution becomes uniform. Figure 5 shows the pdf for problem instances with $|\text{COMPS}| = 100$ for an increasing fault cardinality $|\omega|$. In order to decrease sampling noise, the empirical $f(k)$ values in Fig. 5 are computed by taking the average over 10 samples of $k$.

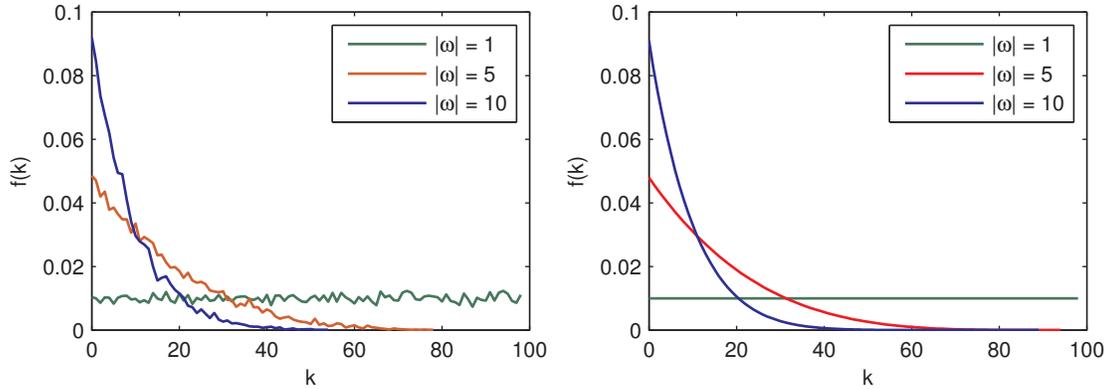

Figure 5: Empirical (left) and analytic (right) $f(k)$ for no retries and a single diagnosis

In the next section we show that retries will further move probability mass towards the optimum, increasing the tail of the distribution, which is needed for (almost always) reaching optimality.

### 4.1.2 Modeling Retries

In this section we extend the model to account for retries, which has a profound effect on the resulting pdf of $f$. Again, consider the transition between step $k$ and $k + 1$, where the algorithm can spend up to $m = 1, \ldots, M$ retries before exiting with failure. As can be seen by the algorithm (cf. Alg. 1), when a variable flip produces an inconsistency a retry is executed while $m$ is incremented.

From elementary combinatorics we can compute the probability of having a diagnosis after flipping any of $M$ different negative literals at step $k$. Similar to (3), at stage $k$ there are $|\mathrm{COMPS}| - k$ faulty literals from which $M$ are chosen (as variable "flips" leading to inconsistency are recorded and not attempted again, there is no difference between choosing the $M$ variables in advance or one after another). The probability of advancing from stage $k$ to stage $k+1$ becomes:

$$p'(k) = 1 - \frac{\binom{|\omega|}{M}}{\binom{|\mathrm{COMPS}|-k}{M}} \tag{6}$$

The progress of Safari can be modeled for values of $M > 1$ as a Markov chain, similar to the one shown in Fig. 4 with the transition probability of $p$ replaced by $p'$. The resulting pdf of the number of successful steps becomes:

$$f'(k) = \frac{\binom{|\omega|}{M}}{\binom{|\mathrm{COMPS}|-k+1}{M}} \prod_{i=0}^{k} \left[ 1 - \frac{\binom{|\omega|}{M}}{\binom{|\mathrm{COMPS}|-i}{M}} \right] \tag{7}$$

It can be seen that (5) is a restricted case of (7) for $M = 1$.

The retry effect on the shape of the pdf is profound. Whereas for single-fault solutions the shape for $M = 0$ is uniform, for $M = 1$ most of the probability mass is already located at the optimum $k = |\mathrm{COMPS}| - |\omega|$. Fig. 6 plots $f$ for a number of problem instances with increasing $M$. As expected, the effect of $M$ is extremely significant. Note that in case of the real system, for $M = |\mathrm{COMPS}|$ the pdf would consist of a single, unit probability spike at $|\mathrm{COMPS}| - |\omega|$.

Although we were unable to find an analytic treatment of the transition model above, the graphs immediately show that for large $M$ the probability of moving to $k = |\mathrm{COMPS}| - |\omega|$ is very large. Hence, we expect the pdf to have a considerable probability mass located at $k = |\mathrm{COMPS}| - |\omega|$, depending on $M$ relative to $|\mathrm{COMPS}|$.

## 4.2 Optimality of Safari in Strong-Fault Models

From the above analysis we have seen that in **WFM** it is easy, starting from a non-minimal diagnosis, to reach a subset minimal diagnosis. As will be discussed in more detail below, this is not necessarily the case for strong-fault models. In many practical cases, however, strong-fault models exhibit, at least partially, behavior similar to MDH, thus allowing greedy algorithms like Safari to achieve results that are close to the optimal values.

### 4.2.1 Partial Continuity in Strong-Fault Stuck-At Models

In what follows we will restrict our attention to a large subclass of **SFM**, called **SFSM** (Struss & Dressler, 1992).

**Definition 14** (Strong-Fault Stuck-At Model). A system DS = $\langle \mathrm{SD}, \mathrm{COMPS}, \mathrm{OBS} \rangle$ belongs to the class **SFSM** iff SD is equivalent to $(h_1 \Rightarrow F_1) \wedge (\neg h_1 \Rightarrow l_1) \wedge \cdots \wedge (h_n \Rightarrow F_n) \wedge (\neg h_n \Rightarrow l_n)$ such that $1 \leq i, j \leq n$, $\{h_i\} \subseteq \mathrm{COMPS}$, $F_j$ is a propositional formula, none of $h_i$ appears in $F_j$, and $l_j$ is a positive or negative literal in $F_j$.

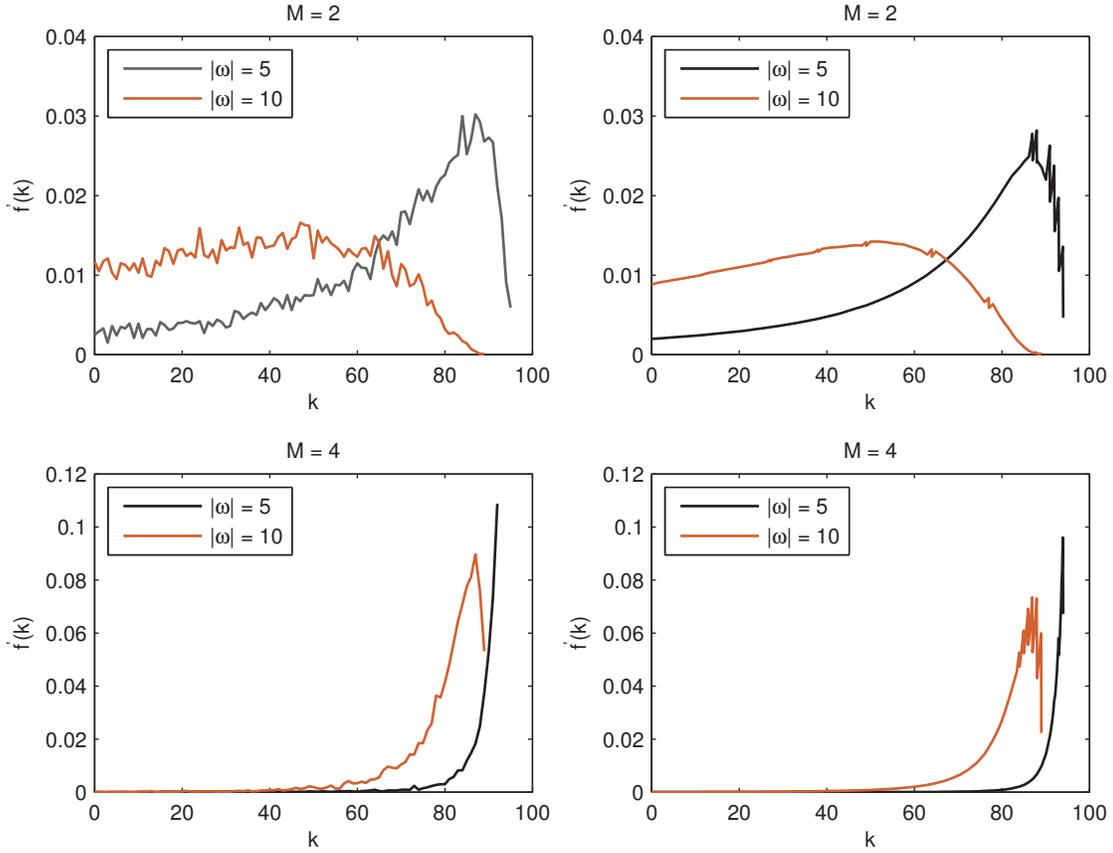

Figure 6: Empirical (left) and analytic (right) $f'(k)$ for multiple retries and a single diagnosis

MDH (cf. Hypothesis 1) does not hold for **SFSM** models. Consider an adder whose inputs and outputs are all zeroes, and whose gate models are all stuck-at-1 when faulty. In this case, the "all nominal" assignment is a diagnosis, but, for example, a stuck-at-1 output gate is not a diagnosis (there is a contradiction with the zero output).

Many practical observations involving **SFSM** models, however, lead to partial continuity. This means that there are groups of diagnoses that differ in at most one literal, i.e., a flip based search can improve the cardinality of a diagnosis. We next formalize this notion.

**Definition 15** (Partial Continuity). A system model SD and an observation $\alpha$ satisfy the partial continuity property with respect to a set $S \subset \Omega(\text{SD} \wedge \alpha)$, iff for every diagnosis $\omega$ such that $\exists \omega_i \in S$ satisfying $Lit^-(\omega) \setminus Lit^-(\omega_i)$ there exists a finite sequence of valid subset flips from $\omega_i$ to $\omega$.

At one extreme of the spectrum, SD and $\alpha$ satisfy the partial continuity property with respect to the set of all of its diagnoses while at the other extreme, the partial continuity property is satisfied with respect to a singleton $S$ (consider, for example, SD $\in$ **WFM** where $S$ consists of the single "all faulty" diagnosis).

Note that the continuous property is trivally satisfied with respect to any diagnosis $\omega_k \in \Omega(\text{SD} \wedge \alpha)$, i.e., there always exists a sequence containing $\omega_k$ only $(\Phi = \langle \omega_k \rangle)$. We are only interested in the non-trivial cases, for which $|\Phi| > 1$.

Consider a system SD and an observation $\alpha$ that satisfy the partial continuity property with respect to some diagnosis $\omega_k$. We say that the diagnoses in the flip sequence $\Phi$ that contains $\omega_k$ form a continuous subspace. Alternatively, given a diagnostic system SD and an observation $\alpha$, a continuous diagnostic subspace of SD$\wedge\alpha$ is a set of diagnoses $\bar{\Omega} \subseteq \Omega(\text{SD} \wedge \alpha)$ with the property that, for any diagnosis $\omega \in \bar{\Omega}$, there is another diagnosis $\bar{\omega} \in \bar{\Omega}$ such that $|Lit^-(\omega)| - |Lit^-(\bar{\omega})| = \pm 1$.

Unfortunately, in the general **SFSM** case, we cannot derive bounds for the sizes of the continuous subspaces, and hence, for the optimality of SAFARI. In what follows, and with the help of a few examples, we illustrate the fact that partial continuity depends on the model and the observation and then we express the optimality of SAFARI as a function of this topologically-dependent property. Later, in Sec. 6, we collect empirical data that continuous subspaces leading to near-optimal diagnoses exist for a class of benchmark **SFSM** circuits.

Our first example illustrates the notion of discontinuity (lack of partial continuity with respect to any diagnoses). We show a rare example of a model and and an observation leading to a set of diagnoses that contains diagnoses of cardinality $m$ and $m + q$ ($q > 1$), but has no diagnoses of cardinality $m + 1, m + 2, \cdots, m + q - 1$.

**A Discontinuity Example**   Consider, for example, the Boolean circuit shown in Fig. 7 and modeled by the propositional formula:

$$\text{SD}_d = \left\{ \begin{array}{l} [h_1 \Rightarrow (y \Leftrightarrow \neg x)] \wedge [\neg h_1 \Rightarrow (y \Leftrightarrow x)] \\ [h_2 \Rightarrow (y \Leftrightarrow \neg x)] \wedge [\neg h_2 \Rightarrow (y \Leftrightarrow x)] \end{array} \right. \tag{8}$$

and an observation $\alpha_d = x \wedge \neg y$. Note, that SD$_d \notin$ **SFSM**. There are exactly two diagnoses of SD$_d \wedge \alpha_d$: $\omega_{15} = h_1 \wedge h_2$ and $\omega_{16} = \neg h_1 \wedge \neg h_2$. Note that this model cannot have single faults. As only $\omega_{15}$ is minimal, $|\omega_{15}| = 0$, and $|\omega_{16}| = 2$, if the algorithm starts from $\omega_{16}$ it is not possible to reach the minimal diagnosis $\omega_{15}$ by performing single flips. Similarly we can construct models which impose an arbitrarily bad bound on the optimality of SAFARI. Such models, however, are not common and we will see that the greedy algorithm performs well on a wide class of strong-fault models.

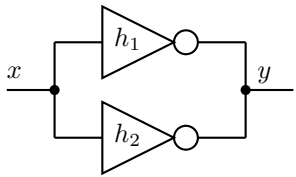

Figure 7: A two inverters circuit

Obviously, continuity in the distribution of the cardinalities in a set of diagnoses is a necessary (but not sufficient) condition for SAFARI to progress. Such models impose arbitrary difficulty to SAFARI, leading to suboptimal diagnoses of any cardinality.

**An Example of Partial Continuity**  We continue the running example started in Sec. 2. First, we create a system description $SD_{sa}$ for a **SFSM** model. Let $SD_{sa} = SD_w \wedge SD_f$, where $SD_w$ is given by (1). The second part of $SD_{sa}$, the strong fault description $SD_f$, specifies that the output of a faulty gate must be stuck-at-1:

$$\begin{aligned} SD_f = (\neg h_1 \Rightarrow i) \wedge (\neg h_2 \Rightarrow d) \wedge (\neg h_3 \Rightarrow j) \wedge (\neg h_4 \Rightarrow m) \wedge \\ \wedge (\neg h_5 \Rightarrow b) \wedge (\neg h_6 \Rightarrow l) \wedge (\neg h_7 \Rightarrow k) \end{aligned} \quad (9)$$

It is clear that $SD_{sa} \in \mathbf{SFSM}$. We next compute the diagnoses of $SD_{sa} \wedge \alpha_1$ ($\alpha_1 = x \wedge y \wedge p \wedge b \wedge \neg d$). There is one minimal diagnosis of $SD_{sa} \wedge \alpha_1$ and it is $\omega_5^{\subseteq} = \neg h_1 \wedge h_2 \wedge h_3 \wedge \cdots \wedge h_7$ (cf. Fig. 8). If we choose the two literals $h_3$ and $h_4$ from $\omega_5^{\subseteq}$ and change the signs of $h_3$ and $h_4$, we create two new health assignments: $\omega_{15} = \neg h_1 \wedge h_2 \wedge \neg h_3 \wedge h_4 \wedge h_5 \wedge h_6 \wedge h_7$ and $\omega_{16} = \neg h_1 \wedge h_2 \wedge h_3 \wedge \neg h_4 \wedge h_5 \wedge h_6 \wedge h_7$. It can be checked that both $\omega_{15}$ and $\omega_{16}$ are diagnoses, i.e., $SD_{sa} \wedge \alpha_1 \wedge \omega_{15} \not\models \perp$ and $SD_{sa} \wedge \alpha_1 \wedge \omega_{16} \not\models \perp$. Note that $\omega_{15}$ and $\omega_{16}$ are diagnoses of the weak-part of the model, i.e., $\{\omega_{15}, \omega_{16}\} \subset \Omega(SD_w \wedge \alpha_1)$. This follows from MDH and the fact that $\omega_5^{\subseteq}$ is a minimal diagnosis of $SD_w \wedge \alpha_1$. Furthermore, $\omega_{15}$ is also a diagnosis in the strong-fault stuck-at model ($\omega_{15} \in \Omega(SD_{sa} \wedge \alpha_1)$) because $SD_w \wedge \alpha_1 \wedge \neg h_3$ does not lead to a contradictory value for $j$ in the strong-fault part $SD_f$. A similar argument applies to $\omega_{16}$: $SD_w \wedge \alpha_1 \wedge \neg h_4$ does not contradict $m$ in $SD_f$. Equivalently, if negating $h_3$ in $\omega_5^{\subseteq}$, which makes $j$ stuck-at-1, results in a diagnosis, and negating $h_4$ in $\omega_5^{\subseteq}$, which makes $m$ stuck-at-1, also results in a diagnosis, negating *both* $h_3$ and $h_4$ in $\omega_5^{\subseteq}$ will also result in a diagnosis (consider the fact that the fault mode of $h_4$ sets $m$ only, but does not impose constraints on $j$). The above argument can be extended similarly to $h_5$, $h_6$, and $h_7$. Hence, any assignment of COMPS containing $\neg h_1 \wedge h_2$ is a diagnosis of $SD_{sa} \wedge \alpha_1$, no matter what combination of signs we take for $h_3$, $h_4$, $h_5$, $h_6$, and $h_7$. Note that a health assignment containing $\neg h_4$ is a diagnosis conditioned on $k = 1$.

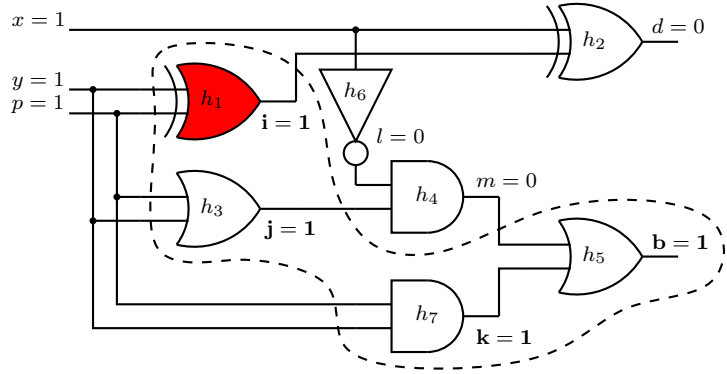

Figure 8: Continuous subspace in a strong-fault, stuck-at-1 model of a subtractor

Consider an alternative way of computing a set of ambiguous diagnoses of $SD_{sa} \wedge \alpha_1$. Given $SD_{sa} \wedge \alpha_1 \wedge \omega_5^{\subseteq}$, we can compute a consistent assignment to all internal variables (for example by propagation). There is exactly one such assignment $\phi$ and it is $\phi = i \wedge j \wedge k \wedge \neg l \wedge \neg m$, $SD_{sa} \wedge \alpha_1 \wedge \omega_5^{\subseteq} \wedge \phi \not\models \perp$ (cf. Fig. 8). Note that for components $h_1$, $h_3$, $h_5$, and $h_7$, a change in the state of a component (healthy or faulty) does not lead to a different output value. For example the output $j$ of the $h_3$ or-gate is 1 because the gate is healthy and its inputs are 1 but $j$ would also be 1 for a stuck-at-1 or-gate ($\neg h_3$). As a result, no diagnostic reasoner can determine if the components in the dashed region of Fig. 8 are healthy or faulty (stuck-at-1). Equivalently, one can change the signs of $h_3$, $h_5$, and $h_7$ in the diagnosis $\omega_5^\subseteq$ and the resulting assignments are still diagnoses. We call the set of components modeled by $h_1$, $h_3$, $h_5$, and $h_7$ an ambiguity group. Clearly, SAFARI can start from a diagnosis $\omega_{17} = \neg h_1 \wedge h_2 \wedge \neg h_3 \wedge h_4 \wedge \neg h_5 \wedge h_6 \wedge \neg h_7$ ($|\omega_{17}| = 4$) and reach $\omega_5^\subseteq$ ($|\omega_5^\subseteq| = 1$) by performing valid subset flips.

To make our reasoning precise, we restrict the class of **SFSM** models to exclude malformed circuits such as ones having disconnected inputs or outputs, etc. Furthermore, we assume that each component has exactly one output (the set of all component output variables is denoted as COUT). The latter is not a big restriction as multi-output component models can be replaced by multiple components, each having a single output.[6]

**Definition 16** (Well-Formed Diagnostic System (**Wfds**)). The diagnostic system DS = $\langle$SD, COMPS, OBS$\rangle$ is well-formed (DS $\in$ **Wfds**) iff for any observation $\alpha$ and for any diagnosis $\omega \in \Omega(\text{SD} \wedge \alpha)$, there is exactly one assignment $\phi$ to all component outputs COUT such that SD $\wedge \alpha \wedge \omega \wedge \phi \not\models \perp$.

Consider an **SFSM** model SD = $(h_1 \Rightarrow F_1) \wedge (\neg h_1 \Rightarrow l_1) \wedge \cdots \wedge (h_n \Rightarrow F_n) \wedge (\neg h_n \Rightarrow l_n)$. We denote as COMPS$^-$ the set of those $h_i$ ($1 \leq i \leq n$) for which the respective $l_i$ literals are negative (cf. Def. 14), i.e., COMPS$^-$ is the set of components whose failure modes are stuck-at-0. Similarly, we use COMPS$^+$ for the set of component variables whose stuck-at $l_i$ literals are positive (COMPS$^- \cup$ COMPS$^+$ = COMPS, COMPS$^- \cap$ COMPS$^+$ = $\emptyset$). In a **Wfds**, an observation $\alpha$ and a diagnosis $\omega$ force the output of each component either to a negative or to a positive value. We denote the set of health variables whose respective component outputs are forced to negative values as $G^-(\text{DS}, \alpha, \omega)$. Similarly, we have $G^+(\text{DS}, \alpha, \omega)$ for the components whose outputs have positive values. With all this we can define the notion of a component ambiguity group.

**Definition 17** (Component Ambiguity Group). Given a system DS = $\langle$SD, COMPS, OBS$\rangle$, SD $\in$ **SFSM**, SD $\in$ **Wfds**, an observation $\alpha$, and a diagnosis $\omega \in \Omega(\text{SD} \wedge \alpha)$, the component ambiguity group $\mathcal{U}(\text{DS}, \alpha, \omega)$, $\mathcal{U} \subseteq$ COMPS, is defined as $\mathcal{U}(\text{DS}, \alpha, \omega) = \{G^-(\text{DS}, \alpha, \omega) \cap$ COMPS$^-\} \cup \{G^+(\text{DS}, \alpha, \omega) \cap$ COMPS$^+\}$.

Finally, we show that a component ambiguity group leads to a continuous subspace. In the general case we cannot say much about the size of the component ambiguity groups. From experimentation, we have noticed that it is difficult to assign the inputs of an **SFSM** to values that generate small continuous subspaces (either SD $\wedge \alpha \models \perp$, or SD $\wedge \alpha$ leads to large component ambiguity groups). Of course, it is possible to consider an adder, or a multiplier, for example, whose inputs are all zeroes and whose gate models are all stuck-at-1 when faulty, but the number of such inputs/circuit combinations is small.

**Proposition 2.** *A diagnostic system SD, SD $\in$ **SFSM**, SD $\in$ **Wfds**, and an observation $\alpha$ entail continuous diagnostic subspaces.*

---

6. Any multi-output Boolean function can be replaced by a composition of single-output Boolean functions.

*Proof.* From Def. 16 and the fact that SD ∈ **Wfds** it follows that the output values of a subset of the components have the same sign as the model's stuck-at value. We denote this set as COMPS′, COMPS′ ⊆ COMPS. Any health assignment $\bar{\omega}$ that differs only in signs of components belonging to COMPS′ is also a diagnosis. If the set of diagnoses of SD ∧ α contains all possible assignments to the assumables in COMPS′ then those diagnoses form a continuous space (cf. Def. 17).  □

To best illustrate Proposition 2, consider the or-gate modeled by $h_3$ in Fig. 8. Its output is 1 either because the gate is healthy and one of the gate's inputs is 1, or because the gate is stuck-at-1. In this situation, it is not possible to determine if the component is healthy or faulty.

Clearly, $|\mathcal{U}(\text{DS}, \alpha, \omega)|$ is a lower bound for the progress of SAFARI in stuck-at models. It can be shown that if SAFARI starts from a diagnosis $\omega$ of maximum cardinality for the given subspace, SAFARI is guaranteed (for $M = |\text{COMPS}|$) to improve the cardinality of $\omega$ by at least $|\mathcal{U}(\text{DS}, \alpha, \omega)|$. In practice, SAFARI can proceed even further as the stuck-at ambiguity groups are only one factor of diagnostic uncertainty. A stuck-at component effectively "disconnects" inputs from outputs, hence gates from the fan-in region are not constrained. For instance, continuing our example, for $\neg h_5$, all predecessors in the cone of $\neg h_5$ (components $\neg h_3$, $\neg h_4$, $\neg h_5$, $\neg h_6$, and $\neg h_7$) constitute a continuous health subspace. Contrary to a component ambiguity group, this set is conditional on the health state of another component. A thorough study of stuck-at continuity is outside the scope of this paper but as we shall see in Sec. 6, continuous subspaces justify SAFARI experiments on stuck-at models.

### 4.2.2 PERFORMANCE MODELING WITH STUCK-AT MODELS

To further study the optimality of SAFARI in strong-fault models, we first define a case in which the algorithm cannot improve a non-minimal diagnosis by changing the sign of a faulty literal. Note that the existence of such cases is not a sufficient condition for SAFARI to be suboptimal, as it is possible to reach a minimal diagnosis by first changing the sign of some other faulty literal, thus "circumventing" the missing diagnosis.

From the preceding section we know that the number of "invalid flips" does not depend on $k$, i.e., it is determined by the observation vector and the fault modes. The probability of SAFARI to progress from any non-minimal diagnosis becomes

$$p(k) = 1 - \frac{\binom{|\omega| + |X|}{M}}{\binom{|\text{COMPS}| - k}{M}} \tag{10}$$

where $|X|$ is the number of "invalid flips". The ratio of the number of "invalid flips" $|X|$ to $|\text{COMPS}|$ we will call **SFM** density $d$. The density $d$ gives the average probability of trying an "invalid flip" throughout the diagnostic search. An approximation of the probability of success of SAFARI is:

$$p(k) = 1 - \frac{\binom{|\omega|}{M}}{\binom{|\text{COMPS}| - k}{M}} - d \tag{11}$$

Plugging $p$ into (4) allows us to predict $f(k)$ for the **SFM** models for which our assumptions hold. This pdf, both measured from an implementation of Safari and generated from (4) and (11) is shown in Fig. 9 for different values of the density $d$.

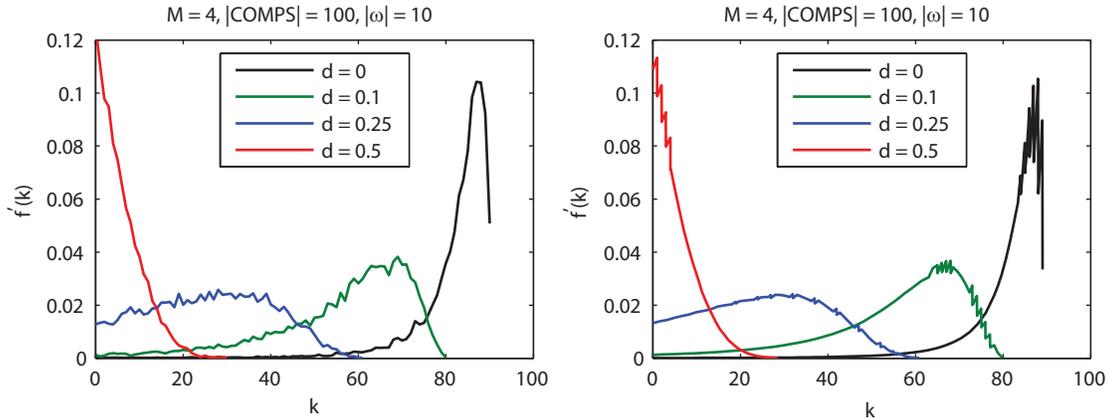

Figure 9: Empirical (left) and analytic (right) $f'(k)$ for various diagnostic densities, multiple retries and a single diagnosis

From Fig. 9 it is visible that increasing the density $d$ leads to a shift of the probability density of the length of the walk $k$ to the left. The effect, however, is not that profound even for large values of $d$, and is easily compensated by increasing $M$, as discussed in the preceding sections.

It is interesting to note that bounds on $d$ can be computed from SD (independent of $\alpha$), and these bounds can be used to further improve the performance of Safari.

## 4.3 Validation

In the preceding sections we have illustrated the progress of Safari with synthetic circuits exposing specific behavior (diagnoses). In the remainder of this section we will plot the pdf of the greedy search on one of the small benchmark circuits (for more information on the 74181 model cf. Sec. 6).

The progress of Safari with a weak-fault model of the 74181 circuit is shown in Fig. 10. We have chosen a difficult observation leading to a minimal diagnosis of cardinality 7 (left) and an easy observation leading to a single fault diagnosis (right). Both plots show that the probability mass shifts to the right when increasing $M$ and the effect is more profound for the smaller cardinality.

The effect of the stuck-at-0 and stuck-at-1 fault modes (**SFM**) on the probability of success of Safari is shown in Fig. 11.

Obviously, in this case the effect of increasing $M$ is smaller, although still depending on the difficulty of the observation vector. Last, even for small values of $M$, the absolute probability of Safari finding a minimal diagnosis is sizeable, allowing the use of Safari as a practical *anytime* algorithm which always returns a diagnosis, the optimality of which depends on the time allocated to its computation.

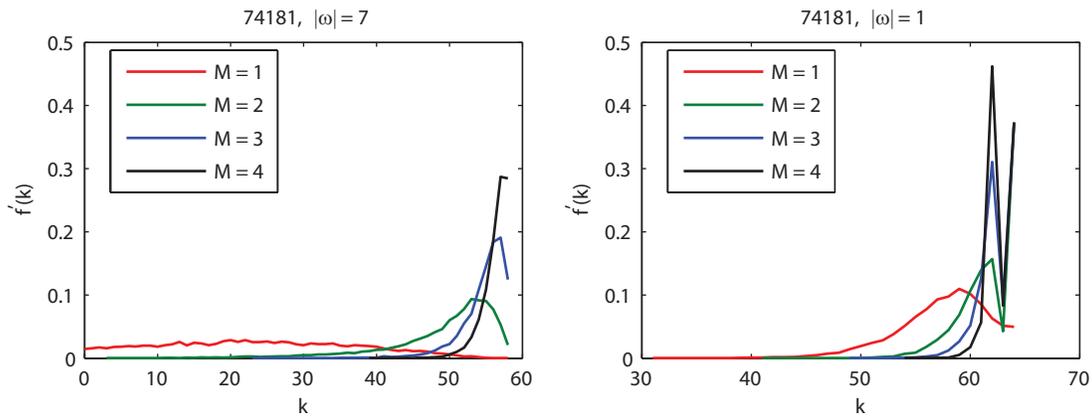

Figure 10: Empirical $f'(k)$ for a weak-fault model of the 74181 circuit with observations leading to two different minimal-cardinality diagnoses and various $M$

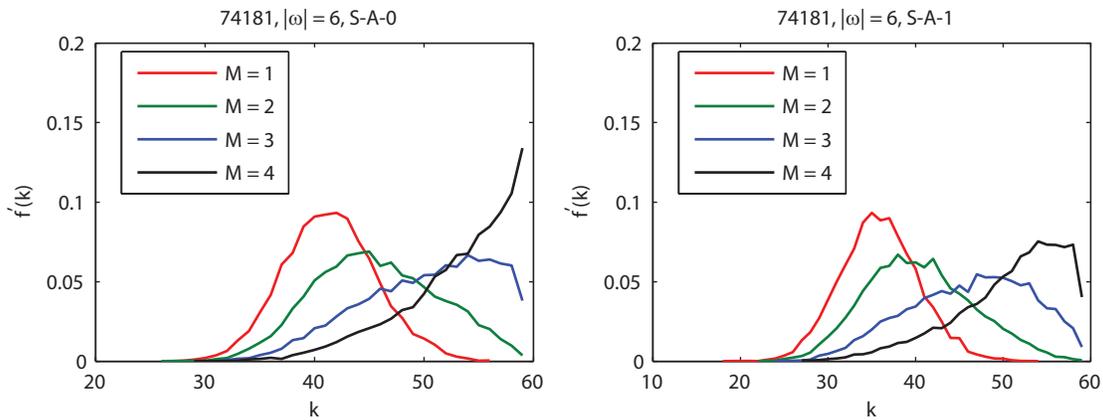

Figure 11: Empirical $f'(k)$ for stuck-at-0 and stuck-at-1 strong-fault models of the 74181 circuit with various $M$

## 5. Optimality Analysis (Multiple Diagnoses)

The preceding section described the process of computing one diagnosis with SAFARI ($N = 1$). In this section we discuss the use of SAFARI in computing (or counting) all minimal-cardinality diagnoses ($N > 1$). For the rest of the section we will assume that SAFARI is configured with $M = |\text{COMPS}|$.

Consider a system description SD (SD $\in$ **WFM**) and an observation $\alpha$. The number of minimal diagnoses $|\Omega^{\subseteq}(\text{SD} \wedge \alpha)|$ can be exponential in $|\text{COMPS}|$. Furthermore, in practice, diagnosticians are interested in sampling from the set of minimal-cardinality diagnoses $\Omega^{\leq}(\text{SD} \wedge \alpha)$ (recall that $\Omega^{\leq}(\text{SD} \wedge \alpha) \subseteq \Omega^{\subseteq}(\text{SD} \wedge \alpha)$) as the minimal-cardinality diagnoses cover a significant part of the *a posteriori* diagnosis probability space (de Kleer, 1990). In what follows, we will see that SAFARI is very well suited for that task.

**Theorem 3.** *The probability of* SAFARI *configured with $M = |\text{COMPS}|$ computing a minimal diagnosis of cardinality $|\omega|$ in a system with $|\text{COMPS}|$ component variables approaches* $|\text{COMPS}|^{-|\omega|}$ *for* $|\text{COMPS}|/|\omega| \to \infty$.

*Proof (Sketch).* Assume a minimal diagnosis of cardinality $|\omega|$ exists. From Proposition 1 it follows that SAFARI configured with $M = |\text{COMPS}|$ is guaranteed to compute minimal diagnoses. Starting from the "all faulty" assignment, consider a step $k$ in "improving" the diagnosis cardinality. If state $k$ contains more than one diagnosis, then at state $k+1$, SAFARI will either (1) flip a literal belonging to this diagnosis (note that a literal may belong to more than one diagnosis) and subsequently prevent SAFARI from reaching this diagnosis or (2) flip a literal belonging to a diagnosis which has already been invalidated (i.e., one or more of its literals have been flipped at an earlier step).

The probability that a solution of cardinality $|\omega|$ "survives" a flip at iteration $k$ (i.e., is not invalidated) is:

$$p(k) = 1 - \frac{|\omega|}{|\text{COMPS}| - k} = \frac{|\text{COMPS}| - |\omega| - k}{|\text{COMPS}| - k} \tag{12}$$

Similarly to our basic model (Sec. 4.1.1), the probability that a diagnosis $\omega$ "survives" until it is returned by the algorithm:

$$f(|\text{COMPS}| - |\omega| - 1) = \prod_{i=0}^{|\text{COMPS}|-|\omega|-1} p(i) = \prod_{i=0}^{|\text{COMPS}|-|\omega|-1} \frac{|\text{COMPS}| - |\omega| - i}{|\text{COMPS}| - i} \tag{13}$$

Rewriting the right hand side of Eq. (13) gives us:

$$f(|\text{COMPS}| - |\omega| - 1) = \frac{(|\text{COMPS}| - |\omega|)!}{(|\omega| + 1)(|\omega| + 2) \cdots |\text{COMPS}|} = \frac{|\omega|!(|\text{COMPS}| - |\omega|)!}{|\text{COMPS}|!} \tag{14}$$

Since

$$\frac{(|\text{COMPS}| - |\omega|)!}{|\text{COMPS}|!} = \frac{1}{(|\text{COMPS}| - |\omega| + 1)(|\text{COMPS}| - |\omega| + 2) \cdots |\text{COMPS}|} \tag{15}$$

it holds that

$$\lim_{|\text{COMPS}|/|\omega| \to \infty} \frac{(|\text{COMPS}| - |\omega|)!}{|\text{COMPS}|!} = |\text{COMPS}|^{-|\omega|} \tag{16}$$

As a result, for small $|\omega|$ relative to $|\text{COMPS}|$,

$$f(|\text{COMPS}| - |\omega| - 1) = |\omega|!|\text{COMPS}|^{-|\omega|} \tag{17}$$

which gives us the above theorem. $\qquad \square$

The distribution $h_i(|\omega|)$ of the cardinalities of the minimal diagnoses in $\Omega^{\subseteq}(SD \wedge \alpha)$ depends on the topology of SD and on $\alpha$; i.e., we can create SD and $\alpha$ having any $h_i(|\omega|)$. We denote the cardinality distribution of the minimal diagnoses computed by SAFARI as $h(|\omega|)$.

Theorem 3 gives us a termination criterion for SAFARI which can be used for enumerating and counting minimal-cardinality diagnoses. Instead of running SAFARI with a fixed $N$, it is sufficient to compute the area under the output distribution function $\sum h$. This value will converge to a single value, hence we can terminate SAFARI after the change of $\sum h$ drops below a fixed threshold. Note that SAFARI is efficient in enumerating the minimal-cardinality diagnoses, as they are computed with a probability that is exponentially higher than that of the probability of computing minimal diagnoses of higher-cardinality, as shown in Theorem 3.

**Corollary 1.** SAFARI *computes diagnoses of equal cardinality with equal probability.*

*Proof (Sketch).* From Theorem 3 it follows that the probability of success $f$ of SAFARI in computing a diagnosis $\omega$ depends only on $|\omega|$ and not on the actual composition of $\omega$. $\quad\square$

The above corollary gives us a simple termination criterion for SAFARI in the cases when all minimal diagnoses are also minimal-cardinality diagnoses; it can be proven that in this case all minimal-cardinality diagnoses are computed with the same probability.

We will see that, given an input cardinality distribution $h_i(|\omega|)$, SAFARI produces an output distribution $h(|\omega|)$ that is highly skewed to the right, due to Theorem 3. To facilitate the study of how SAFARI transforms $h_i(|\omega|)$ into $h(|\omega|)$ we will use a Monte Carlo simulation of SAFARI. The advantage is that the Monte Carlo simulation is much simpler for analysing the run-time behavior of SAFARI than studying the algorithm itself.

---

**Algorithm 2** Monte Carlo simulation of SAFARI

1: **function** SAFARISIMULATE($\Omega^{\subseteq}, N$) **returns** a cardinality distribution
    **inputs:** $\Omega^{\subseteq}$, a set of minimal diagnoses
          $N$, integer, number of tries
    **local variables:** $h_i, h$, vectors, cardinality distributions
              $b$, vector, fault distribution, $n, i, c$, integers
2:    $h_i \leftarrow$ CARDINALITYDISTRIBUTION($\Omega^{\subseteq}$)
3:    **for** $n \leftarrow 1, 2, \dots, N$ **do**
4:       **for** $c \leftarrow 1, 2, \dots, |h_i|$ **do**
5:          $b[c] \leftarrow c \cdot h_i[c]$
6:       **end for**
7:       **for** $i \leftarrow 1, 2, \dots, |\Omega^{\subseteq}|$ **do**
8:          $c \leftarrow$ DISCRETEINVERSERANDOMVALUE $\left( \frac{b}{\sum b} \right)$
9:          $b[c] \leftarrow b[c] - c$
10:      **end for**
11:      $h[c] \leftarrow h[c] + 1$
12:    **end for**
13:    **return** $h$
14: **end function**

---

Algorithm 2 simulates which diagnoses from the input set of minimal diagnoses $\Omega$ are "reached" by Safari in $N$ tries. The auxiliary subroutine CardinalityDistribution computes the input distribution $h_i$ by iterating over all diagnoses in $\Omega^{\subseteq}$. We store the input cardinality distribution $h_i$ and the resulting cardinality distribution $h$ in vectors (note the vector sums in lines 7 and 8 and the division of a vector by scalar in line 8).

The outermost loop of Alg. 2 (lines 3 – 12) simulates the $N$ runs of Safari. This is done by computing and updating an auxiliary vector $b$, which contains the distribution of the component variables in $\Omega^{\subseteq}$ according to the cardinalities of the diagnoses these variables belong to. Initially, $b$ is initialized with the number of literals in single faults in position 1, the number of literals in double faults in position 2 (for example if there are three double faults in $h_i$, $b[2] = 6$), etc. This is done in lines 4 – 6 of Alg. 2. We assume that diagnoses do not share literals. This restriction can be easily dropped by counting all the assumables in the input $\Omega^{\subseteq}$ (the latter assumption does not change the results of this section).

Lines 7 – 10 simulate the process of the actual bit flipping of Safari. At each step the simulation draws a random literal from the probability distribution function ($pdf$) $\frac{b}{\sum b}$; this is done by the DiscreteInverseRandomValue function in line 8. Each bit flip "invalidates" a diagnosis from the set $\Omega^{\subseteq}$, i.e., a diagnosis of cardinality $c$ cannot be reached by Safari. After a diagnosis has been "invalidated", the vector $b$ is updated, for example, if the simulation "invalidates" a quadruple fault, $b[4] = b[4] - 4$ (line 9). Note that the number of iterations in the loop in lines 7 – 10 equals the number of diagnoses in $\Omega^{\subseteq}$. As a result after terminating this loop, the value of the integer variable $c$ is equal to the cardinality of the *last* "invalidated" diagnosis. The latter is the diagnosis which Safari computes in this run. What remains is to update the resulting pdf with the right cardinality (line 11).

The simulation in Alg. 2 links the distribution of the actual diagnoses in $\Omega^{\subseteq}$ to the distribution of the cardinalities of the diagnoses returned by Safari. As $\Omega^{\subseteq}$ can be arbitrarily set, we will apply Alg. 2 to a range of typical input distributions. The results of the simulation as well as the results of running Safari on synthetic problems with the same input distributions are shown in Fig. 12.

Fig. 12 shows (1) that Alg. 2 predicts the actual behavior of Safari (compare the second and third column of plots), and (2) that Safari computes diagnoses of small cardinality in agreement with Theorem 3. The only case when the output distribution is not a steep exponential is when the cardinalities in the set of the input minimal diagnoses grow exponentially. Table 2 summarizes the parameters of exponential fits for the input cardinality distributions shown in Fig. 12 ($a$ is the initial (zero) cardinality, $\lambda$ is the decay constant, and $R^2$ is the coefficient of determination). We have seen that Safari is suited for computing multiple diagnoses of small probability of occurrence. In the next section we will provide an alternative argument leading to similar conclusions.

## 6. Experimental Results

This section discusses empirical results measured from an implementation of Safari. In order to compare the optimality and performance of Safari to various diagnostic algorithms, we have performed more than a million diagnosis computations on 64 dual-CPU nodes belonging to a cluster. Each node contains two 2.4 GHz AMD Opteron DP 250 processors and 4 Gb of RAM.

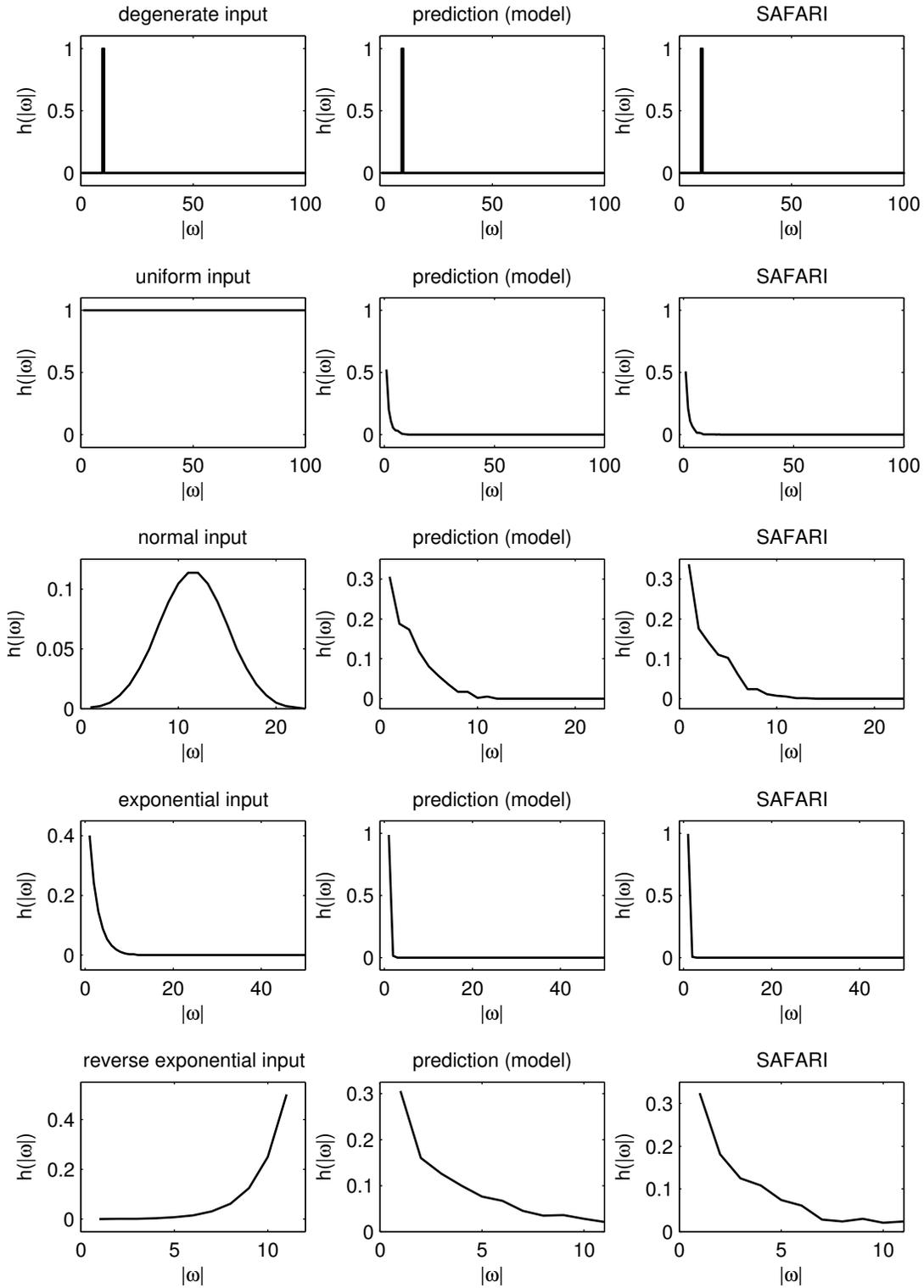

Figure 12: Predicted and actual cardinality distributions

Table 2: Fit coefficients to exponential and goodness of fit for the cardinality distribution in Fig. 12

| Input Distribution | $a$ | $\lambda$ | $R^2$ |
|---|---|---|---|
| Uniform | 576 | $-0.44$ | 1 |
| Normal | 423 | $-0.34$ | 0.99 |
| Exponential | 69 470 | $-4.26$ | 1 |
| Reverse Exponential | 385 | $-0.33$ | 0.95 |

The default configuration of SAFARI (when not stated otherwise) was $M = 8$ and $N = 4$; that is, SAFARI is configured for a maximum number of 8 retries before giving up the climb, and a total of 4 attempts. To provide more precise average run-time optimality and performance data, all stochastic algorithms (i.e., ones based on SLS Max-SAT and SAFARI) have been repeatedly run 10 times on each model and observation vector.

## 6.1 Implementation Notes and Test Set Description

We have implemented SAFARI in approximately 1 000 lines of C code (excluding the LTMS, interface, and DPLL code) and it is a part of the LYDIA package.[7]

Traditionally, MBD algorithms have been tested on diagnostic models of digital circuits like the ones included in the ISCAS85 benchmark suite (Brglez & Fujiwara, 1985). As models derived from ISCAS85 are large (from a traditional diagnostic perspective), we have also considered four medium-sized circuits from the 74XXX family (Hansen, Yalcin, & Hayes, 1999). In order to provide both weak- and strong-fault cases, we have translated each circuit to a weak, stuck-at-0 (S-A-0), and stuck-at-1 (S-A-1) model. In the stuck-at models, the output of each faulty gate is assumed to be the same constant (cf. Def. 14).

The performance of diagnostic algorithms depends to various degrees on the observation vectors (algorithm designers strive to produce algorithms, the performance of which is not dependent on the observation vectors). Hence, we have performed our experimentation with a number of different observations for each model. We have implemented an algorithm (Alg. 3) that generates observations leading to diagnoses of different minimal-cardinality, varying from 1 to nearly the maximum for the respective circuits (for the 74XXX models it is the maximum). The experiments omit nominal scenarios as they are trivial from the viewpoint of MBD.

Algorithm 3 uses a number of auxiliary functions. RANDOMINPUTS (line 3) assigns uniformly distributed random values to each input in IN (note that for the generation of observation vectors we partition the observable variables OBS into inputs IN and outputs OUT and use the input/output information which comes with the original 74XXX/ISCAS85 circuits for simulation). Given the "all healthy" health assignment and the diagnostic system, COMPUTENOMINALOUTPUTS (line 4) performs simulation by propagating the input assignment $\alpha$. The result is an assignment $\beta$ which contains values for each output variable in OUT.

---

7. LYDIA, SAFARI, and the diagnostic benchmark can be downloaded from `http://fdir.org/lydia/`.

---

**Algorithm 3** Algorithm for generation of observation vectors

1: **function** MAKEALPHAS(DS, $N$, $K$) **returns** a set of observations
     **inputs:** DS $= \langle$SD, COMPS, OBS$\rangle$, diagnostic system
              OBS $=$ IN $\cup$ OUT, IN $\cap$ OUT $= \emptyset$
              $N$, integer, number of tries for SAFARI
              $K$, integer, maximal number of diagnoses per cardinality
     **local variables:** $\alpha, \beta, \alpha_n, \omega$, terms
                     $c$, integer, best cardinality so far
                     $A$, set of terms (observation vectors), result
2:    **for** $k \leftarrow 1, 2, \ldots, K$ **do**
3:       $\alpha \leftarrow$ RANDOMINPUTS(IN)
4:       $\beta \leftarrow$ COMPUTENOMINALOUTPUTS(DS, $\alpha$)
5:       $c \leftarrow 0$
6:       **for all** $v \in$ OUT **do**
7:          $\alpha_n \leftarrow \alpha \wedge$ FLIP($\beta, v$)
8:          $\omega \leftarrow$ SMALLESTCARDINALITYDIAGNOSIS(SAFARI(SD, $\alpha_n$, |COMPS|, $N$))
9:          **if** $|\omega| > c$ **then**
10:            $c \leftarrow |\omega|$
11:            $A \leftarrow A \cup \alpha_n$
12:          **end if**
13:       **end for**
14:    **end for**
15:    **return** $A$
16: **end function**

---

The loop in lines 6 – 13 increases the cardinality by greedily flipping the values of the output variables. For each new candidate observation $\alpha_n$, Alg. 3 uses the diagnostic oracle SAFARI to compute a minimal diagnosis of cardinality $c$. As SAFARI returns more than one diagnosis (up to $N$), we use SMALLESTCARDINALITYDIAGNOSIS to choose the one of smallest cardinality. If the cardinality $c$ of this diagnosis increases in comparison to the previous iteration, the observation is added to the list.

By running Alg. 3 we get up to $K$ observations leading to faults of cardinality $1, 2, \ldots, m$, where $m$ is the cardinality of the MFMC diagnosis (Feldman, Provan, & van Gemund, 2008b) for the respective circuit. Alg. 3 clearly shows a bootstrapping problem. In order to create potentially "difficult" observations for SAFARI, we require SAFARI to solve those "difficult" observations. Although we have seen in Sec. 5 that SAFARI is heavily biased towards generating diagnoses of small cardinality, there is no guarantee. To alleviate this problem, for the generation of observation vectors, we have configured SAFARI to compute subset-minimal diagnoses with $M = $ |COMPS| and $N$ increased to 20.

Table 3 provides an overview of the fault diagnosis benchmark used for our experiments. The third and fourth columns show the number of observable and assumable variables, which characterize the size of the circuits. The next three columns show the number of observation vectors with which we have tested the weak, S-A-0, and S-A-1 models. For the stuck-at models, we have chosen those weak-fault model observations which are consistent with their respective system descriptions (as in strong-fault models it is often the case that $SD \wedge \alpha \models \perp$, we have not considered such scenarios).

Table 3: An overview of the 74XXX/ISCAS85 benchmark circuits

| Name | Description | Variables | | Observations | | |
|---|---|---|---|---|---|---|
| | | \|OBS\| | \|COMPS\| | Weak | S-A-0 | S-A-1 |
| 74182 | 4-bit carry-lookahead generator | 14 | 19 | 250 | 150 | 82 |
| 74L85 | 4-bit magnitude comparator | 14 | 33 | 150 | 58 | 89 |
| 74283 | 4-bit adder | 14 | 36 | 202 | 202 | 202 |
| 74181 | 4-bit ALU | 22 | 65 | 350 | 143 | 213 |
| c432 | 27-channel interrupt controller | 43 | 160 | 301 | 301 | 301 |
| c499 | 32-bit SEC circuit | 73 | 202 | 835 | 235 | 835 |
| c880 | 8-bit ALU | 86 | 383 | 1 182 | 217 | 335 |
| c1355 | 32-bit SEC circuit | 73 | 546 | 836 | 836 | 836 |
| c1908 | 16-bit SEC/DEC | 58 | 880 | 846 | 846 | 846 |
| c2670 | 12-bit ALU | 373 | 1 193 | 1 162 | 134 | 123 |
| c3540 | 8-bit ALU | 72 | 1 669 | 756 | 625 | 743 |
| c5315 | 9-bit ALU | 301 | 2 307 | 2 038 | 158 | 228 |
| c6288 | 32-bit multiplier | 64 | 2 416 | 404 | 274 | 366 |
| c7552 | 32-bit adder | 315 | 3 512 | 1 557 | 255 | 233 |

## 6.2 Comparison to Complete Algorithms

Table 4 shows the results from comparing SAFARI to implementations of two state-of-the-art complete and deterministic diagnostic algorithms: a modification for completeness of CDA* (Williams & Ragno, 2007) and HA* (Feldman & van Gemund, 2006). Table 4 shows, for each model and for each algorithm, the percentage of all tests for which a diagnosis could be computed within a cut-off time of 1 minute.

As it is visible from the three rightmost columns of Table 4, SAFARI could find diagnoses for all observation vectors, while the performance of the two deterministic algorithms (columns two to seven) degraded with the increase of the model size and the cardinality of the observation vector. Furthermore, we have observed a degradation of the performance of CDA* and HA* with increased cardinality of the minimal-cardinality diagnoses, while the performance of SAFARI remained unaffected.

## 6.3 Comparison to Algorithms Based on ALLSAT and Model Counting

We have compared the performance of SAFARI to that of a pure SAT-based approach, which uses blocking clauses for avoiding duplicate diagnoses (Jin, Han, & Somenzi, 2005). Although SAT encodings have worked efficiently on a variety of other domains, such as planning, the weak health modeling makes the diagnostic problem so underconstrained that an uninformed ALLSAT strategy (i.e., a search not exploiting the continuity imposed by the weak-fault modeling) is quite inefficient, even for small models.

Table 4: Comparison of CDA*, HA*, and SAFARI [% of tests solved]

| Name | CDA* | | | HA* | | | SAFARI | | |
|------|------|------|------|------|------|------|------|------|------|
|      | Weak | S-A-0 | S-A-1 | Weak | S-A-0 | S-A-1 | Weak | S-A-0 | S-A-1 |
| 74182 | 100 | 100 | 100 | 100 | 100 | 100 | 100 | 100 | 100 |
| 74L85 | 100 | 100 | 100 | 100 | 100 | 100 | 100 | 100 | 100 |
| 74283 | 100 | 100 | 100 | 100 | 100 | 100 | 100 | 100 | 100 |
| 74181 | 79.1 | 98.6 | 97.7 | 100 | 100 | 100 | 100 | 100 | 100 |
| c432 | 74.1 | 75.4 | 73.1 | 71.1 | 94.7 | 69.1 | 100 | 100 | 100 |
| c499 | 29 | 45.5 | 27.7 | 24.1 | 77.9 | 25.9 | 100 | 100 | 100 |
| c880 | 11.6 | 44.7 | 32.2 | 12.4 | 62.2 | 41.5 | 100 | 100 | 100 |
| c1355 | 3.8 | 4.7 | 5.4 | 10.8 | 10.6 | 12.2 | 100 | 100 | 100 |
| c1908 | 0 | 0 | 0 | 6.1 | 6 | 6.5 | 100 | 100 | 100 |
| c2670 | 0 | 0 | 0 | 5 | 64.2 | 44.7 | 100 | 100 | 100 |
| c3540 | 0 | 0 | 0 | 1.1 | 3.8 | 2.2 | 100 | 100 | 100 |
| c5315 | 0 | 0 | 0 | 1.1 | 8.2 | 5.7 | 100 | 100 | 100 |
| c6288 | 0 | 0 | 0 | 3.5 | 5.1 | 3.3 | 100 | 100 | 100 |
| c7552 | 0 | 0 | 0 | 3.9 | 7.8 | 12 | 100 | 100 | 100 |

To substantiate our claim, we have experimented with the state-of-the-art satisfiability solver RELSAT, version 2.02 (Bayardo & Pehoushek, 2000). Instead of enumerating all solutions and filtering the minimal diagnoses only, we have performed model-counting, whose relation to MBD has been extensively studied (Kumar, 2002). While it was possible to solve the two smallest circuits, the solver did not terminate for any of the larger models within the predetermined time of 1 hour. The results are shown in Table 5.

The second column of Table 5 shows the model count returned by RELSAT, with sample single-fault observations from our benchmark. The third column reports the time for model counting. This slow performance on relatively small diagnostic instances leads us to the conclusion that specialized solvers like SAFARI are better suited for finding minimal diagnoses than off-the-shelf ALLSAT (model counting) implementations that do not encode inference properties similar to those encoded in SAFARI.

We have used the state-of-the-art, non-exact model counting method SAMPLECOUNT (Gomes, Hoffmann, Sabharwal, & Selman, 2007) to compute lower bounds of the model counts. The results are shown in the third and fourth columns of Table 5. Configured with the default settings ($\alpha = 3.5$, $t = 2$, $z = 20$, cutoff $10\,000$ flips), SAMPLECOUNT could not find lower bounds for circuits larger than c1355. Although the performance of SAMPLE-COUNT is significantly better than RELSAT, the fact that SAMPLECOUNT computes lower bounds and does not scale to large circuits prevent us from building a diagnosis algorithm based on approximate model counting.

A satisfiability-based method for diagnosing an optimized version of ISCAS85 has been used by Smith, Veneris, and Viglas (2004). In a more recent paper (Smith, Veneris, Ali, & Viglas, 2005), the SAT-based approach has been replaced by a Quantified Boolean Formula (QBF) solver for computing multiple-fault diagnoses. These methods report good absolute execution time for single and double-faults (and we believe that they scale well for higher cardinalities), but require modifications of the initial circuits (i.e., introduce cardinality and test constraints) and suggest specialized heuristics for the SAT solvers in order to improve the search performance. Comparison of the performance of SAFARI to the timings reported by these papers would be difficult due to a number of reasons like the use of different and optimized benchmark sets, trading-off memory for speed, rewriting the original circuits, etc.

Table 5: Model count and time for counting

| Name | RelSat | | SampleCount | |
|------|--------|--------|-------------|--------|
| | Models | Time [s] | Models | Time [s] |
| 74182 | $3.9896 \times 10^{7}$ | 1 | $\geq 3.526359 \times 10^{6}$ | 0.2 |
| 74L85 | $8.3861 \times 10^{14}$ | 340 | $\geq 7.412344 \times 10^{13}$ | 0.3 |
| 74283 | $\geq 1.0326 \times 10^{15}$ | $> 3\,600$ | $\geq 3.050026 \times 10^{14}$ | 0.3 |
| 74181 | $\geq 5.6283 \times 10^{15}$ | $> 3\,600$ | $\geq 1.538589 \times 10^{27}$ | 1.1 |
| c432 | $\geq 7.2045 \times 10^{18}$ | $> 3\,600$ | $\geq 1.496602 \times 10^{67}$ | 9.9 |
| c499 | $\geq 3.6731 \times 10^{20}$ | $> 3\,600$ | $\geq 7.549183 \times 10^{83}$ | 13.1 |
| c880 | $\geq 9.4737 \times 10^{39}$ | $> 3\,600$ | $\geq 8.332702 \times 10^{166}$ | 42.7 |
| c1355 | $\geq 1.4668 \times 10^{28}$ | $> 3\,600$ | $\geq 7.488300 \times 10^{233}$ | 99.8 |
| c1908 | $\geq 2.1704 \times 10^{31}$ | $> 3\,600$ | — | — |
| c2670 | $\geq 9.0845 \times 10^{15}$ | $> 3\,600$ | — | — |
| c3540 | $\geq 4.8611 \times 10^{19}$ | $> 3\,600$ | — | — |
| c5315 | $\geq 9.3551 \times 10^{16}$ | $> 3\,600$ | — | — |
| c6288 | $\geq 1.0300 \times 10^{18}$ | $> 3\,600$ | — | — |
| c7552 | $\geq 1.0049 \times 10^{16}$ | $> 3\,600$ | — | — |

## 6.4 Performance of the Greedy Stochastic Search

Table 6 shows the absolute performance of SAFARI ($M = |\text{COMPS}|, N = 4$). This varies from under a millisecond for the small models, to approx. 30 s for the largest strong-fault model. These fast absolute times show that SAFARI is suitable for on-line reasoning tasks, where autonomy depends on speedy computation of diagnoses.

For each model, the minimum and maximum time for computing a diagnosis has been computed. These values are shown under columns $t_{min}$ and $t_{max}$, respectively. The small range of $t_{max} - t_{min}$ confirms our theoretical results that SAFARI is insensitive to the fault cardinalities of the diagnoses it computes. The performance of CDA* and HA*, on the other hand, is dependent on the fault cardinality and quickly degrades with increasing fault cardinality.

## 6.5 Optimality of the Greedy Stochastic Search

From the results produced by the complete diagnostic methods (CDA* and HA*) we know the exact cardinalities of the minimal-cardinality diagnoses for some of the observations. By considering these observations, which lead to single and double faults, we have evaluated the average optimality of SAFARI. Table 7 shows these optimality results for the greedy search. The second column of Table 7 shows the number of observation vectors leading to single faults for each weak-fault model. The third column shows the average cardinality of SAFARI. The second and third column are repeated for the S-A-0 and S-A-1 models.

Table 6: Performance of SAFARI [ms]

| Name | Weak | | S-A-0 | | S-A-1 | |
|------|------|------|------|------|------|------|
| | $t_{\min}$ | $t_{\max}$ | $t_{\min}$ | $t_{\max}$ | $t_{\min}$ | $t_{\max}$ |
| 74182 | 0.41 | 1.25 | 0.39 | 4.41 | 0.40 | 0.98 |
| 74L85 | 0.78 | 7.47 | 0.72 | 1.89 | 0.69 | 4.77 |
| 74283 | 0.92 | 4.84 | 0.88 | 3.65 | 0.92 | 5.2 |
| 74181 | 2.04 | 6.94 | 2.13 | 22.4 | 2.07 | 7.19 |
| c432 | 8.65 | 38.94 | 7.58 | 30.59 | 7.96 | 38.27 |
| c499 | 14.19 | 31.78 | 11.03 | 30.32 | 10.79 | 31.11 |
| c880 | 48.08 | 88.87 | 37.08 | 80.74 | 38.47 | 81.34 |
| c1355 | 95.03 | 141.59 | 76.57 | 150.29 | 83.14 | 135.29 |
| c1908 | 237.77 | 349.96 | 196.13 | 300.11 | 217.32 | 442.91 |
| c2670 | 500.54 | 801.12 | 646.95 | 1 776.72 | 463.24 | 931.8 |
| c3540 | 984.31 | 1 300.98 | 1 248.5 | 2 516.46 | 976.56 | 2 565.18 |
| c5315 | 1 950.12 | 2 635.71 | 3 346.49 | 7 845.41 | 2 034.5 | 4 671.17 |
| c6288 | 2 105.28 | 2 688.34 | 2 246.84 | 3 554.4 | 1 799.18 | 2 469.48 |
| c7552 | 4 557.4 | 6 545.21 | 9 975.04 | 32 210.71 | 5 338.97 | 12 101.61 |

Table 7 shows that, for SD $\in$ **WFM**, the average cardinality returned by SAFARI is near-optimal for both single and double faults. The c1355 model shows the worst-case results for the single-fault observations, while c499 is the most difficult weak-fault model for computing a double-fault diagnosis. These results can be improved by increasing $M$ and $N$ as discussed in Sec. 4.

With strong-fault models, results are close to optimal for the small models and the quality of diagnosis deteriorates for c3540 and bigger. This is not surprising, considering the modest number of retries and number of "flips" with which SAFARI was configured.

## 6.6 Computing Multiple Minimal-Cardinality Diagnoses

We next show the results of experiments supporting the claims made in Sec. 5. For that, we have first chosen these observations $\alpha$ for which we could compute $|\Omega^{\leq}(\text{SD} \wedge \alpha)|$ with a deterministic algorithm like CDA* or HA* (mostly observations leading to single or double faults). We have then configured SAFARI with $M = |\text{COMPS}|$ and $N = 10|\Omega^{\leq}(\text{SD} \wedge \alpha)|$. Finally, from the diagnoses computed by SAFARI we have filtered the minimal-cardinality ones. The results are summarized in Table 8.

Table 8 repeats the same columns for weak, S-A-0, and S-A-1 models and the data in these columns are to be interpreted as follows. The columns marked with $|\Omega^{\leq}|$ show the minimal and maximal number of minimal-cardinality diagnoses per model as computed by a deterministic algorithm. The columns $M_c$ show the percentage of minimal-cardinality diagnoses returned by SAFARI (from all minimal-cardinality diagnoses) for those $\alpha$ for which $|\Omega^{\leq}(\text{SD} \wedge \alpha)| > 1$. The columns $M_f$ show the percentage of observations for which SAFARI could not compute any minimal-cardinality diagnosis.

Table 7: Optimality of SAFARI [average cardinality]

| | Single Faults | | | | | | Double Faults | | | | | |
| | Weak | | S-A-0 | | S-A-1 | | Weak | | S-A-0 | | S-A-1 | |
| Name | # | Card. | # | Card. | # | Card. | # | Card. | # | Card. | # | Card. |
|---|---|---|---|---|---|---|---|---|---|---|---|---|
| 74182 | 50 | 1 | 37 | 1 | 40 | 1 | 50 | 2 | 38 | 2 | 18 | 2 |
| 74L85 | 50 | 1.04 | 18 | 1.02 | 40 | 1.03 | 50 | 2.12 | 17 | 2.06 | 35 | 2.07 |
| 74283 | 50 | 1.08 | 34 | 1.59 | 46 | 1.88 | 50 | 2.2 | 45 | 2.41 | 42 | 2.6 |
| 74181 | 50 | 1.19 | 36 | 2.81 | 46 | 2.6 | 50 | 2.25 | 36 | 3.61 | 43 | 3.16 |
| c432 | 58 | 1.19 | 52 | 1.06 | 37 | 1.04 | 82 | 2.46 | 80 | 2.25 | 48 | 2.15 |
| c499 | 84 | 1.49 | 53 | 1.49 | 84 | 1.01 | 115 | 3.27 | 34 | 3.01 | 115 | 2.03 |
| c880 | 50 | 1 | 39 | 1.1 | 40 | 1.05 | 50 | 2.01 | 34 | 2.14 | 35 | 2.07 |
| c1355 | 84 | 1.66 | 82 | 1 | 84 | 1.02 | 6 | 2.15 | 7 | 2 | 18 | 2.07 |
| c1908 | 52 | 1.05 | 49 | 2.91 | 52 | 4.79 | — | — | 2 | 3 | 3 | 3.17 |
| c2670 | 29 | 1.03 | 39 | 1.77 | 28 | 2.06 | 13 | 2.12 | 24 | 2.78 | 15 | 3.27 |
| c3540 | 8 | 1.01 | 23 | 2.5 | 16 | 3.74 | — | — | 1 | 4.9 | — | — |
| c5315 | 14 | 1 | 9 | 3.54 | 12 | 5.4 | 7 | 2 | 3 | 3.7 | 1 | 3.8 |
| c6288 | 13 | 1 | 13 | 28.83 | 12 | 28.68 | 1 | 2 | 1 | 27 | — | — |
| c7552 | 27 | 1.01 | 11 | 17.37 | 18 | 23.38 | 16 | 2 | 4 | 18.5 | 6 | 27.53 |

Table 8: % of all minimal-cardinality diagnoses computed by SAFARI

| | Weak | | | S-A-0 | | | S-A-1 | | |
| Name | $|\Omega^{\leq}|$ | $M_c$ | $M_f$ | $|\Omega^{\leq}|$ | $M_c$ | $M_f$ | $|\Omega^{\leq}|$ | $M_c$ | $M_f$ |
|---|---|---|---|---|---|---|---|---|---|
| 74182 | $1-25$ | 100 | 0 | $1-2$ | 100 | 0 | $1-20$ | 100 | 0 |
| 74L85 | $1-78$ | 99.2 | 2 | $1-4$ | 100 | 0 | $1-49$ | 99.7 | 0 |
| 74283 | $1-48$ | 97.9 | 3 | $1-16$ | 93.8 | 0 | $1-29$ | 84.9 | 4 |
| 74181 | $1-133$ | 97.4 | 1 | $1-16$ | 88.6 | 4.07 | $1-57$ | 96.7 | 6.36 |
| c432 | $1-99$ | 94.2 | 7.14 | $1-40$ | 89.7 | 0 | $1-18$ | 97 | 0 |
| c499 | $1-22$ | 78.5 | 1.51 | $1-15$ | 96.3 | 6 | $1-16$ | 94.8 | 0 |
| c880 | $2-646$ | 99.9 | 0 | $1-160$ | 96.9 | 0 | $1-210$ | 97.5 | 0 |
| c1355 | $5-2\,770$ | 79.4 | 1.02 | $2-648$ | 95.7 | 0 | $2-347$ | 95.2 | 0.52 |
| c1908 | $2-1\,447$ | 96.6 | 2.61 | $2-579$ | 85.2 | 1.85 | $2-374$ | 82.3 | 1.24 |
| c2670 | $1-76$ | 100 | 2.34 | $1-20$ | 97.1 | 0 | $1-181$ | 89.7 | 0 |
| c3540 | $1-384$ | 81.5 | 8.52 | $1-153$ | 88.8 | 7.98 | $1-171$ | 78.2 | 7.27 |
| c5315 | $1-235$ | 97.7 | 1.74 | $1-24$ | 81.7 | 7.04 | $1-30$ | 93.4 | 8.24 |
| c6288 | $1-154$ | 100 | 13.1 | $1-73$ | 78.1 | 5.1 | $1-101$ | 82.1 | 1.22 |
| c7552 | $1-490$ | 93.1 | 2.17 | $4-236$ | 90.8 | 13.55 | $1-168$ | 78 | 12.1 |

The results shown in Table 8 show that even for moderate values of $N$ ($N \leq 27\,770$), SAFARI was capable of computing a significant portion of all minimal-cardinality diagnoses. This portion varies from 78.5% to 100% for weak-fault models and from 78% to 100% for strong-fault models. The percentage of cases in which SAFARI could not reach a minimal-cardinality diagnosis is limited (at most 13.55%) and is mainly in the cases in which there exists only one single-fault diagnosis. Note that even in the cases in which SAFARI cannot compute any minimal-cardinality diagnoses, the result of SAFARI can still be useful. For example, a subset-minimal diagnosis of small cardinality differing in one or two literals only nevertheless brings useful diagnostic information (a discussion on diagnostic metrics is beyond the scope of this paper).

## 6.7 Experimentation Summary

We have applied SAFARI to a suite of benchmark combinatorial circuits encoded using weak-fault models and stuck-at strong fault models, and shown significant performance improvements for multiple-fault diagnoses, compared to two state-of-the-art deterministic algorithms, CDA* and HA*. Our results indicate that SAFARI shows at least an order-of-magnitude speedup over CDA* and HA* for multiple-fault diagnoses. Moreover, whereas the search complexity for the deterministic algorithms tested increases exponentially with fault cardinality, the search complexity for this stochastic algorithm appears to be independent of fault cardinality.

We have compared the performance of SAFARI to that of an algorithm based on Max-SAT, and SAFARI shows at least an order-of-magnitude speedup in computing diagnoses. We have compared the optimality of SAFARI to that of an algorithm based on SLS Max-SAT, and SAFARI consistently computes diagnoses of smaller cardinality whereas the SLS Max-SAT diagnostic algorithm often fails to compute any diagnosis.

## 7. Related Work

This paper (1) generalizes Feldman, Provan, and van Gemund (2008a), (2) introduces important theoretical results for strong-fault models, (3) extends the experimental results there, and (4) provides a comprehensive optimality analysis of SAFARI.

On a gross level, one can classify the types of algorithms that have been applied to solve MBD as being based on search or compilation. The search algorithms take as input the diagnostic model and an observation, and then search for a diagnosis, which may be minimal with respect to some minimality criterion. Examples of search algorithms include A*-based algorithms, such as CDA* (Williams & Ragno, 2007) and hitting set algorithms (Reiter, 1987). Compilation algorithms pre-process the diagnostic model into a form that is more efficient for on-line diagnostic inference. Examples of such algorithms include the ATMS (de Kleer, 1986) and other prime-implicant methods (Kean & Tsiknis, 1993), DNNF (Darwiche, 1998), and OBDD (Bryant, 1992). To our knowledge, all of these approaches adopt exact methods to compute diagnoses; in contrast, SAFARI adopts a stochastic approach to computing diagnoses.

At first glance, it seems like MBD could be efficiently solved using an encoding as a SAT (Jin et al., 2005), constraint satisfaction (Freuder, Dechter, Ginsberg, Selman, & Tsang, 1995) or Bayesian network (Kask & Dechter, 1999) problem. However, one needs to take into account the increase in formula size (over a direct MBD encoding), in addition to the underconstrained nature of MBD problems.

SAFARI has close resemblance to Max-SAT (Hoos & Stützle, 2004) and we have conducted extensive experimentation with both complete (partial and weighted) and SLS-based Max-SAT. As the results of these experiments are long, we have published them in a separate technical report (Feldman, Provan, & van Gemund, 2009a). The results show that although Max-SAT can compute diagnoses in many of the cases, the performance of Max-SAT degrades when increasing the circuit size or the cardinality of the injected faults. In particular, SAFARI outperforms Max-SAT by at least an order-of-magnitude for the class of diagnostic problems we have considered. In the case of SLS-based Max-SAT, the optimality of Max-SAT-based inference is significantly worse than that of SAFARI.

We show that SAFARI exploits a particular property of MBD problems, called diagnostic continuity, which improves the optimality of SAFARI compared to, for example, straightforward ALLSAT encodings (Jin et al., 2005). We experimentally confirm this favorable performance and optimality of SAFARI. Although SAFARI has close resemblance to Max-SAT, SAFARI exploits specific landscape properties of the diagnostic problems, which allow (1) simple termination criteria and (2) optimality bounds. Due to the hybrid nature of SAFARI (the use of LTMS and SAT), SAFARI avoids getting stuck in local optima and performs better than Max-SAT based methods. Incorporating approaches from Max-SAT, and in particular SAPS (Hutter, Tompkins, & Hoos, 2002), in future versions of SAFARI may help in solving more general abduction problems, which may not expose the continuous properties of the models we have considered.

Stochastic algorithms have been discussed in the framework of constraint satisfaction (Freuder et al., 1995) and Bayesian network inference (Kask & Dechter, 1999). The latter two approaches can be used for solving suitably translated MBD problems. It is often the case, though, that these encodings are more difficult for search than specialized ones.

MBD is an instance of constraint optimization, with particular constraints over failure variables. MBD has developed algorithms to exploit these domain properties, and our proposed approach differs significantly from almost all MBD algorithms that appear in the literature. While most advanced MBD algorithms are deterministic, SAFARI borrows from SLS algorithms that, rather than backtracking, may randomly flip variable assignments to determine a satisfying assignment. Complete MBD algorithms typically make use of preferences, e.g., fault-mode probabilities, to improve search efficiency; SAFARI uses this technique on top of its stochastic search over the space of diagnoses.

A closely-related diagnostic approach is that of Fijany, Vatan, Barrett, James, Williams, and Mackey (2003), who map the minimal-hitting set problem into the problem of finding an assignment with bounded weight satisfying a monotone SAT problem, and then propose to use efficient SAT algorithms for computing diagnoses. The approach of Fijany et al. has shown speedups in comparison with other diagnosis algorithms; the main drawback is the number of extra variables and clauses that must be added in the SAT encoding, which is even more significant for strong fault models and multi-valued variables. In contrast, our approach works directly on the given diagnosis model and requires no conversion to another representation.

Our work bears the closest resemblance to preference-based or Cost-Based Abduction (CBA) (Charniak & Shimony, 1994; Santos Jr., 1994). Of the algorithmic work in this area, the primary paper that adopts stochastic local search is by Abdelbar, Gheita, and Amer (2006). In this paper, they present a hybrid two-stage method that is based on Iterated Local Search (ILS) and Repetitive Simulated Annealing (RSA). The ILS stage of the algorithm uses a simple hill-climbing method (randomly flipping assumables) for the local search phase, and tabu search for the perturbation phase. RSA repeatedly applies Simulated Annealing (SA), starting each time from a random initial state. The hybrid method initially starts from an arbitrary state, or a greedily-chosen state. It then applies the ILS algorithm; if this algorithm fails to find the optimal solution after a fixed number $\tau$ of hill-climbing steps[8] or after a fixed number $\mathcal{R}$ of repetitions of the perturbation-local search cycle,[9] ILS-based search is terminated and the RSA algorithm is run until the optimal solution is found.

Our work differs from that of Abdelbar et al. (2006) in several ways. First, our initial state is generated using a random SAT solution. The hill-climbing phase that we use next is similar to that of Abdelbar et al.; however, we randomly restart should hill-climbing not identify a "better" diagnosis, rather than applying tabu search or simulated annealing. Our approach is simpler than that of Abdelbar et al., and for the case of weak fault models is guaranteed to be optimal; in future work we plan to compare our approach to that of Abdelbar et al. for strong fault models.

In 2009 Safari competed against the diagnostic algorithms NGDE (de Kleer, 2009) and RODON (Bunus, Isaksson, Frey, & Münker, 2009) in the synthetic track of the first diagnostic competition DXC'09 (Kurtoglu, Narasimhan, Poll, Garcia, Kuhn, de Kleer, van Gemund, & Feldman, 2009). The conditions under which the DXC'09 experiments were conducted were similar to the ones described in this paper. The CPU and memory performance of Safari were an order of magnitude better than the competing algorithms despite the fact that NGDE and RODON performed better than the complete algorithms discussed in this section. In this paper, in addition to computational metrics, we have informally used the minimality of a diagnosis as an optimality criterion. The DXC'09 organizers, however, have defined a utility metric which approximates the expected repair effort of a circuit (Feldman, Provan, & van Gemund, 2009b). With this utility metric, Safari scored slightly worse than the two competing algorithms, which is to be expected as Safari trades off diagnostic precision for computational efficiency. We refer the reader to the DXC papers mentioned above for a more thorough analysis of the competition results.

## 8. Conclusion and Future Work

We have described a greedy stochastic algorithm for computing diagnoses within a model-based diagnosis framework. We have shown that subset-minimal diagnoses can be computed optimally in weak fault models and in an important subset of strong fault models, and that almost all minimal-cardinality diagnoses can be computed for more general fault models.

---

8. Hill-climbing proceeds as follows: given a current state $s$ with a cost of $f(s)$, a neighbouring state $s'$ is generated by flipping a randomly chosen assumable hypothesis. If $f(s')$ is better than $f(s)$, then $s'$ becomes the current state; otherwise, it is discarded. If $\tau$ iterations elapse without a change in the current state, the local search exits.

9. Perturbation-local search, starting from a current state $s$ with a cost of $f(s)$, randomly chooses an assumable variable $h$, and applies tabu search to identify a better state by flipping $h$ based on its tabu status.

We argue that Safari can be of broad practical significance, as it can compute a significant fraction of minimal-cardinality diagnoses for systems too large or complex to be diagnosed by existing deterministic algorithms.

In future work, we plan to experiment on models with a combination of weak and strong failure-mode descriptions. We also plan on experimenting with a wider variety of stochastic methods, such as simulated annealing and genetic search, using a larger set of benchmark models. Last, we plan to apply our algorithms to a wider class of abduction and constraint optimization problems.